\documentclass[10pt,twocolumn,letterpaper]{article}

\usepackage[pagenumbers]{cvpr} %

\usepackage{multirow}
\usepackage{subcaption}
\usepackage{graphicx}
\usepackage{booktabs}
\usepackage{array}

\definecolor{cvprblue}{rgb}{0.21,0.49,0.74}
\usepackage[pagebackref,breaklinks,colorlinks,allcolors=cvprblue]{hyperref}

\def\paperID{4758} %
\def\confName{CVPR}
\def\confYear{2025}

\title{EigenGS Representation: From Eigenspace to Gaussian Image Space}

\author{
Lo-Wei Tai$^1$ \quad
Ching-En Li$^1$ \quad
Cheng-Lin Chen$^1$ \\[2pt]%
Chih-Jung Tsai$^1$ \quad%
Hwann-Tzong Chen$^{1,2}$ \quad
Tyng-Luh Liu$^3$\\[2pt]
{\small
$^1$National Tsing Hua University\quad
$^2$Aeolus Robotics \quad
$^3$Academia Sinica, Taiwan
}
}

\begin{document}
\maketitle

\begin{abstract}
    Principal Component Analysis (PCA), a classical dimensionality reduction technique, and 2D Gaussian representation, an adaptation of 3D Gaussian Splatting for image representation, offer distinct approaches to modeling visual data. We present EigenGS, a novel method that bridges these paradigms through an efficient transformation pipeline connecting eigenspace and image-space Gaussian representations. Our approach enables instant initialization of Gaussian parameters for new images without requiring per-image optimization from scratch, dramatically accelerating convergence. EigenGS introduces a frequency-aware learning mechanism that encourages Gaussians to adapt to different scales, effectively modeling varied spatial frequencies and preventing artifacts in high-resolution reconstruction. Extensive experiments demonstrate that EigenGS not only achieves superior reconstruction quality compared to direct 2D Gaussian fitting but also reduces necessary parameter count and training time. The results highlight EigenGS's effectiveness and generalization ability across images with varying resolutions and diverse categories, making Gaussian-based image representation both high-quality and viable for real-time applications.
\end{abstract}
    
\section{Introduction}

\begin{figure}[t]
 \centering
  \begin{subfigure}[b]{0.47\textwidth}
    \includegraphics[width=\textwidth]{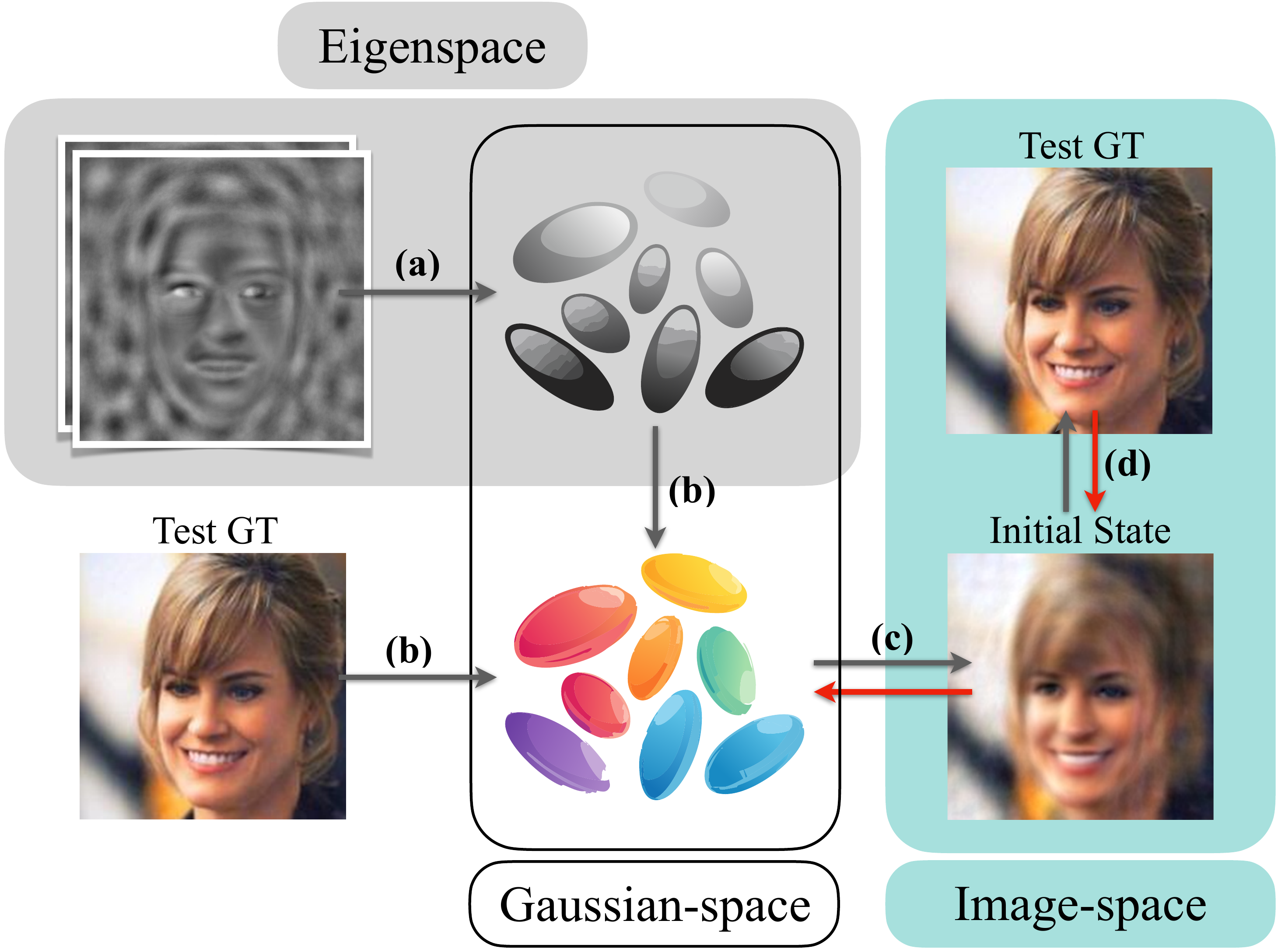}
  \end{subfigure}
  \caption{Overview of our method:  
(a) Our approach learns a set of eigen-Gaussian representations that characterize the principal components in eigenspace.
(b) For a given test image, we can instantly derive its image-based Gaussian representation in image space from the eigen-Gaussians.
(c) The initial image-space Gaussian representation serves as a robust starting point, enabling rapid convergence to reconstruct the input image.  
(d) We further optimize this representation by minimizing the image-space Gaussian reconstruction loss between the rendered output and the test image to obtain a high-quality final result. Here, we use red arrows to represent the gradient flow from the reconstruction loss.}
\label{fig:overview}
\end{figure}

\begin{figure*}
    \centering
    \begin{minipage}{0.24\textwidth}
        \centering
        $\widetilde{I}^{(0)}\,$ (ITER = 0) 
    \end{minipage}%
    \begin{minipage}{0.24\textwidth}
        \centering
        $\widetilde{I}^{(10)}\,$ (ITER = 10)
    \end{minipage}%
    \begin{minipage}{0.24\textwidth}
        \centering
        $\widetilde{I}^{(100)}\,$ (ITER = 100)
    \end{minipage}%
    \begin{minipage}{0.24\textwidth}
        \centering
        GT
    \end{minipage}
    \vspace{0.3em}

    \begin{subfigure}[b]{0.23\textwidth}
        \includegraphics[width=\textwidth]{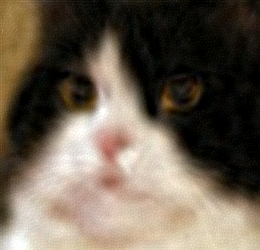}
    \end{subfigure}
    \begin{subfigure}[b]{0.23\textwidth}
        \includegraphics[width=\textwidth]{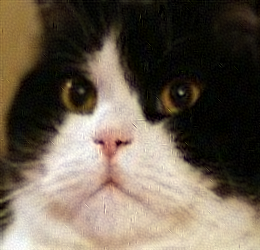}
    \end{subfigure}
    \begin{subfigure}[b]{0.23\textwidth}
        \includegraphics[width=\textwidth]{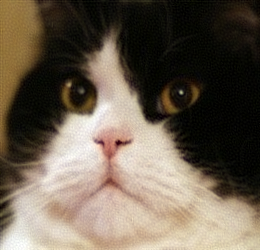}
    \end{subfigure}
    \begin{subfigure}[b]{0.23\textwidth}
        \includegraphics[width=\textwidth]{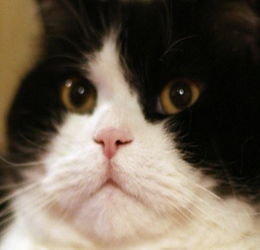}
    \end{subfigure}

    \caption{Fast convergence speed on Cats dataset. Our method achieves perceptually significant improvements within the first 100 iterations. The rapid visual convergence, exhibited by $\widetilde{I}^{(0)} \rightarrow \widetilde{I}^{(10)} \rightarrow \widetilde{I}^{(100)}$, demonstrates the effectiveness of our PCA-based initialization, particularly on well-aligned datasets with moderate resolution. Additionally, our result also indicates that traditional PCA reconstruction, which is basically similar to our result at iteration zero, can be further improved by our approach.}
    \label{fig:convergence}
\end{figure*}

Principal Component Analysis (PCA) is a widely recognized method for dimensionality reduction \cite{Jolliffe86}. It and its variant have been extensively used in computer vision and pattern recognition as a means of achieving low-dimensional representations of visual data \cite{BelhumeurHK97,SirovichK87,TippingB99,TorreB01,TurkP91,Ye04}.
A common approach to applying PCA to images is to treat each pixel as a separate dimension in a vector. This allows for the derivation of a basis of eigenvectors, which can then be used as components to reconstruct an image through linear combination. These components can be visualized as ``eigenimages'' and are useful for transforming an image from an image space to an eigenspace.

Treating each pixel as a separate dimension in a vector may overlook the locality property between neighboring pixels, as well as the self-similarity found among groups of pixels. Although low-frequency and high-frequency patterns can still be implicitly captured from the data, PCA-based image representations do not take advantage of the relationships between nearby or nonlocal pixels in the image.
Significant evidence such as ConvNets and Vision Transformers \cite{LeCunBDHHHJ89, DosovitskiyB0WZ21} has demonstrated the effectiveness of modeling local and nonlocal properties in images.

A recent study by GaussianImage~\cite{ZhangGXHWQLGZ24} utilizes the concept of 3D Gaussian Splatting (3DGS)~\cite{KerblKLD23} and applies it to images by fitting a mixture of Gaussians across the image domain to reconstruct a single image. Unlike the original 3DGS, the rendering process used by GaussianImage is significantly simplified by disregarding the depth ordering of the Gaussians. With the learned Gaussian representation, each Gaussian encompasses a neighborhood around the Gaussian center, and each pixel is represented by multiple Gaussians. Such a representation may relax the constraints of traditional pixel-wise image representation.

Although GaussianImage is effective in reconstructing the target image, a notable drawback is that each image requires individual training to optimize the Gaussian model from scratch. 
In this work, we propose a new method that connects the representations of the eigenspace with those of the Gaussian image space. We show that by using a pre-trained Gaussian model in the eigenspace, we can immediately estimate the Gaussian representation in the image domain for a new input image. Furthermore, our method also introduces an inherent mechanism for learning the scales of Gaussians in relation to the spatial frequencies of images.

\Cref{fig:overview} illustrates the main ideas and advantages of our methods, 2D EigenGS. We propose a novel and effective way to derive Gaussian representations for images. It leverages the strengths of both PCA and Gaussian-based approaches, bridging eigenspaces and Gaussian image spaces. Our method addresses the issue of training from scratch for an individual image and enables efficient estimation of Gaussian representations. We conduct extensive experiments to demonstrate that EigenGS not only achieves superior reconstruction quality but also significantly speeds up the convergence. \Cref{fig:convergence} shows an example of high-quality reconstruction with a rapid convergence speed. Our experimental results also highlight EigenGS's effectiveness and its ability to generalize across images with varying resolutions and diverse categories.

\section{Related Work}
\subsection{Dimensionality Reduction}

Traditional approaches to image representation often rely on dimensionality reduction techniques, with PCA being one of the most fundamental \cite{Jolliffe86}. In computer vision, PCA and its variants \cite{TippingB99, TorreB01} have been extensively applied to face recognition \cite{BelhumeurHK97, SirovichK87, TurkP91}, image compression, and feature extraction. However, with the advent of deep learning approaches in computer vision, these classical dimensionality reduction methods have gradually taken a back seat. 

PCA, though effective at capturing global image statistics and valued for its mathematical elegance through eigendecomposition, shows several notable limitations in modern applications. Its linear structure falls short in modeling the complex, nonlinear patterns often inherent in natural images. Moreover, PCA's assumption of pixel-wise independence disregards critical local spatial relationships and self-similarity patterns, which are essential for high-quality image reconstruction. Combined with the difficulties of managing high-dimensional data, these limitations have made PCA less attractive in the deep learning era, where end-to-end, trainable models now dominate.

\subsection{Gaussian-based Representations}

Advances in 3D Gaussian Splatting \cite{KerblKLD23} have inspired new approaches to image representation. Unlike traditional discrete representations, Gaussian-based methods offer several advantages, such as explicit representation and efficient computation. Also, the original 3D Gaussian Splatting work has demonstrated the effectiveness of optimizable Gaussian primitives for representing 3D scenes, showing the potential of Gaussian-based approaches for high-quality rendering. Following this direction, several approaches have focused on improving various aspects of Gaussian-based methods. 

Recent research has explored reducing storage requirements through vector quantization and hash-grid assistance, making these representations more practical for resource-constrained devices \cite{HAC2024, Compact3D2024, morgenstern2023compact}. Studies such as \cite{MiniSplatting2024, zhang2024pixel, wewer2024latentsplat, GaussReg2024} have investigated different ways to better optimize the Gaussians including novel regularization method \cite{FreGS2023}, multi-scale mechanisms for detail preservation \cite{BAGS2025}, and geometric optimization \cite{li2025geogaussian}. Additionally, there has been significant work focused on refining particular quality attributes, notably in the context of anti-aliasing methodologies \cite{yu2023mipsplattingaliasfree3dgaussian, MultiScaleAntiAliasing2024, liang2024analytic}. 

Notably, GaussianImage \cite{ZhangGXHWQLGZ24} recently adapted these concepts to 2D image representation by removing depth ordering constraints and introducing an accumulated rendering technique. Their method shows that 2D Gaussian primitives could effectively represent images while enabling efficient rendering. This accumulated rendering technique, which eliminates the need for depth-based sorting and alpha blending, forms an essential foundation for our work in combining eigenspace analysis with Gaussian representations.

\section{Our Method}
This section elaborates on the derivation of our proposed method, termed 2D EigenGS. We demonstrate that employing 2D Gaussians to learn the eigenbasis facilitates the effective construction of Gaussian splatting representations within the image domain for novel input images.

\subsection{Revisiting the Eigenbasis}
For the sake of discussion, we consider an image $I(x, y)$ of dimensions $w$ by $h$ as a 2D array of intensity values. The derivation and results can be generalized to color images without loss of generality.

Let $\mathcal{D} = \{I_1, I_2, \ldots, I_m\}$ be the set of training images whose average is calculated by $\Psi_0 = \frac{1}{m} \sum_{i=1}^m I_i$.
We subtract this ``average image'' from each sample in $\mathcal{D}$ to obtain the adjusted images $I'_i = I_i - \Psi_0$. Each adjusted image $I'_i$ is then reshaped into a vector $\mathbf{v}_i'$ of dimensions $d=w \cdot h$. Using Principal Component Analysis (PCA), we compute the eigenvectors of the data covariance matrix \(C\), which is defined as
\begin{equation}
C = \frac{1}{m} \sum_{i=1}^m {\mathbf{v}_i'} {\mathbf{v}_i'}^T \in \mathbb{R}^{d\times d}\,,
\end{equation}
\noindent
where we can obtain a \emph{truncated} eigenbasis by selecting $k$ eigenvectors corresponding to the top-$k$ eigenvalues of $C$ and $k < d$. Since each of these $k$ eigenvectors can be reshaped to a 2D array of size $w$ by $h$,  we refer to them as \emph{eigenimages} $\{\Psi_j\}_{j=1}^k$, where each $\Psi_j$ is a component of the truncated eigenbasis represented as a $w$ by $h$ image.

\subsection{EigenGS: Gaussian Splatting with Eigenbasis}
From the eigenimages $\Psi_j$, we aim to learn a set $\mathcal{N}$, comprising 2D Gaussians parametrized by spatial mean and covariance with visual attributes to represent the truncated eigenbasis. As we shall describe subsequently, the formulation is analogous to the GaussianImage technique by Zhang~\etal~\cite{ZhangGXHWQLGZ24}, which employs depth-invariant Gaussian Splatting in 2D to overfit an RGB image. Our approach intends to learn a unified 2D Gaussian Splatting model that can approximate all the $k$ components of the eigenbasis. 

The rasterization of the Gaussians to approximate the eigen component $\Psi_j$ at pixel position $(x, y)$ is given by 
\begin{equation}
\widetilde{\Psi}_j (x, y) = \sum_{n=1}^{|\mathcal{N}|} \psi_{n, j}' \cdot \exp\left(-\boldsymbol{\sigma}_{n} (x,y) \right) \,,
\label{eq:rasterization}
\end{equation}
where the reduced weight $\psi_{n, j}'$ takes into account the projected opacity and the component value associated with the $n$th Gaussian owing to the orderless nature of depth-invariant 2D Gaussians. The blending factor $\exp\left(-\boldsymbol{\sigma}_{n} (x, y)\right)$ is specified by $\boldsymbol{\sigma}_{n} = \frac{1}{2} \mathbf{d}_n^T \boldsymbol{\Sigma}_n^{-1} \mathbf{d}_n$, where $\boldsymbol{\Sigma}_n$ is the 2-by-2 projected Gaussian covariance, and $\mathbf{d}_n$ is the displacement between the projected Gaussian center and the pixel position $(x, y)$.

According to PCA, given an image $I$ and the truncated eigenbasis $\{\Psi_j\}_{j=1}^k$ calculated from the training set, we can solve for the linear combination coefficients $\{w_j\}_{j=1}^k$ by projecting $I - \Psi_0$ onto each eigenbasis component, and the image can be approximated by $ \Psi_0 + \sum_{j=1}^k w_j \Psi_j$.

Now considering the Gaussian Splatting estimate $\{\widetilde{\Psi}_j\}$ of the truncated eigenbasis, we can integrate the coefficients $\{w_j\}$ with the rasterization process in \cref{eq:rasterization} to facilitate an efficacious initialization for image rendering by
\begin{equation}
\begin{split}
\widetilde{I}^{(0)} &=  \Psi_0 + \sum_{j=1}^k w_j \widetilde{\Psi}_j \\
&=
 \Psi_0 + \sum_{j=1}^k\sum_{n=1}^{|\mathcal{N}|} w_j \psi_{n, j}' \cdot \exp\left(-\boldsymbol{\sigma}_{n}  \right) \\
&=
 \Psi_0 + \sum_{n=1}^{|\mathcal{N}|} \left( \sum_{j=1}^k w_j \psi_{n, j}' \right) \cdot \exp\left(-\boldsymbol{\sigma}_{n}  \right)  \\
 &=
 \Psi_0 + \sum_{n=1}^{|\mathcal{N}|} c'_n \cdot \exp\left(-\boldsymbol{\sigma}_{n}  \right)  \,.
\end{split}
\label{eq:gspca}
\end{equation}

From \cref{eq:gspca}, we see that the same set of $|\mathcal{N}|$ Gaussians learned for the eigenbasis can be seamlessly transformed into an equivalent set of Gaussians for rendering a new image in the ``image domain'' using the reduced weight $c'_n$ associated with the $n$th Gaussian. Moreover, as will be elucidated in the next section and through our experimental results, we can further refine the attributes of the Gaussians to improve the quality of the reconstruction. Suffice it to say that with our EigenGS-powered initialization, the subsequent fine-tuning step to update the 2D Gaussian parameters of $\mathcal{N}$ leads to rapid convergence of rendering $\widetilde{I}^{(t)}$. An example is shown in \Cref{fig:convergence}.

\begin{figure}
    \centering

    \begin{subfigure}[b]{0.44\textwidth}
        \includegraphics[width=\textwidth]{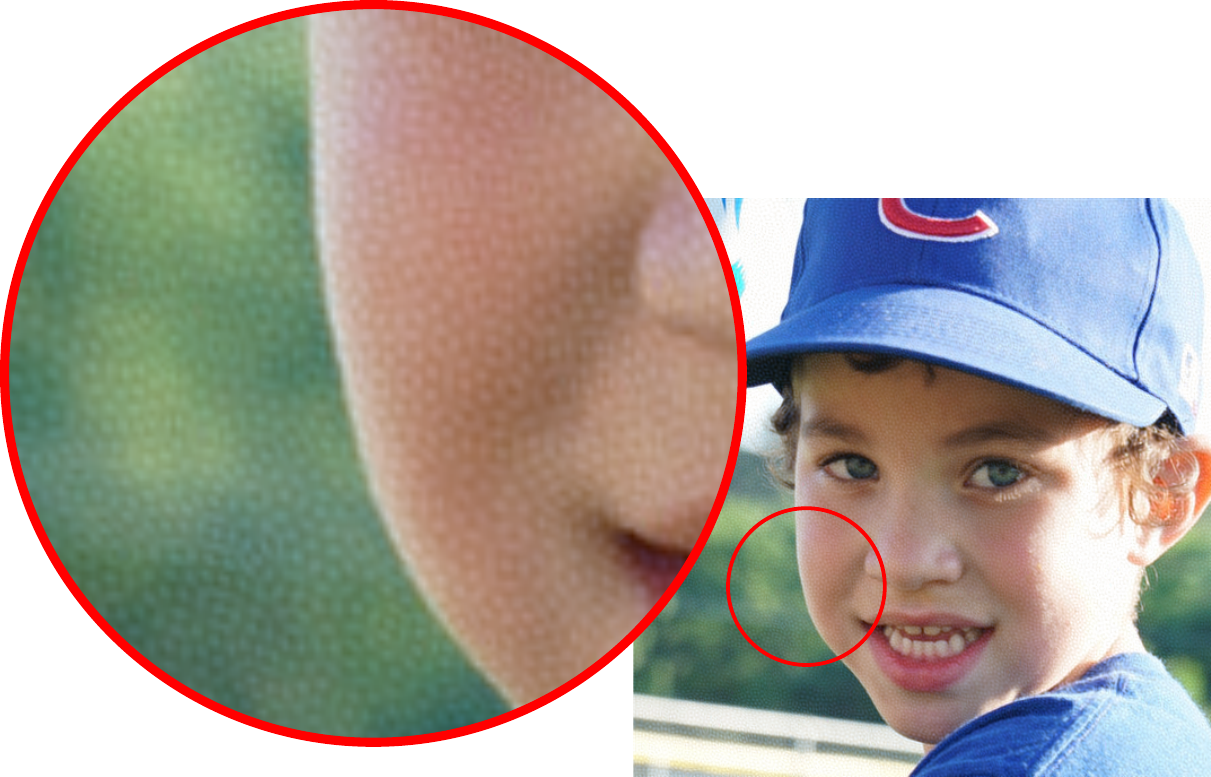}
    \end{subfigure}

    \caption{The ``penny-round-tile" artifact that emerges when Gaussians converge to uniformly small sizes during optimization. Our frequency-aware learning scheme addresses this issue by modeling different spatial frequencies with appropriately sized Gaussians.}
    \label{fig:penny_tile_effect}
\end{figure}

\subsection{Modeling Spatial Frequencies}
Our method effectively mitigates the occurrence of visually distinct artifacts, particularly noticeable as ``penny-round-tile'' artifacts, characterized by small spots surrounding pixels. These artifacts arise due to overly shrunk Gaussians, particularly for high-resolution images, as shown in \Cref{fig:penny_tile_effect}. If the training process proceeds without proper constraints, spatially distributed Gaussians covering the entire image may converge into uniformly small-sized Gaussians to capture high-frequency details and image features during later stages of learning iterations. It is observed that the optimization tends to favor smaller Gaussians when the main objective is to reconstruct the image based on pixel-wise differences. A common technique to prevent overfitting is to incorporate regularization terms associated with the sizes of the Gaussians. Rather than imposing explicit constraints on the sizes of Gaussians, our approach offers an alternative way to regularize the Gaussians so that they can \emph{better} represent the spectrum of spatial frequencies of an image.

In \cref{eq:rasterization}, each of the $|\mathcal{N}|$ Gaussians is formulated to model the entire spectrum of the $k$ eigenbasis components. We instead propose to strategically partition the Gaussians and eigen components into groups such that specific portions of the components are modeled by corresponding portions of the Gaussians. Concretely, we divide all the learnable Gaussians, $\mathcal{N}$, into two disjoint sets, $\mathcal{N}_l$ and $\mathcal{N}_h$, such that $\mathcal{N} = \mathcal{N}_l \dot\cup \mathcal{N}_h$. The $k$ components are likewise divided into two disjoint subsets, $\{\widetilde{\Psi}_l\}$ for the low-frequency components and $\{\widetilde{\Psi}_h\}$ for the high-frequency components, according to their respective eigenvalues in PCA.

We then reformulate \cref{eq:rasterization} such that $\mathcal{N}_l$ and $\mathcal{N}_h$ pertain respectively to ${\widetilde{\Psi}_l}$ and ${\widetilde{\Psi}_h}$. It is clear that the derivation of \cref{eq:gspca} remains valid when using the independently learned Gaussians. 
For those Gaussian weights $\psi_{n, j}'$ with index pairs $(n, j)$ that do not align with any combination in $\mathcal{N}_l \times {\widetilde{\Psi}_l}$ or $\mathcal{N}_h \times {\widetilde{\Psi}_h}$, their values are indeed zero and do not contribute to the final weight $c'_n$. 

In practice, the learning of Gaussians, $\mathcal{N}_l$ and  $\mathcal{N}_h$, can be systematically accomplished in two distinct phases. During the low-frequency phase, we allocate approximately 10\% of the total Gaussians to represent the eigenbasis components corresponding to the larger eigenvalues. While restricting such a relatively small portion of Gaussians to this training phase initially results in higher loss values and slower convergence rates, the model eventually converges as the Gaussians adjust to accommodate larger scales. The subsequent high-frequency stage utilizes the remaining Gaussians to model the components corresponding to the smaller eigenvalues, with a focus on capturing fine details. The final representation combines the outputs from both phases, constituting a mixture of large and small Gaussians that effectively captures both frequency ranges. This useful scheme reduces the ``penny-round-tile" artifact, which can adversely affect the visual quality of the reconstruction.

\section{Experiment}

\begin{table*}
\centering
\resizebox{0.98\linewidth}{!}{%
\setlength{\tabcolsep}{2pt}
\begin{tabular}{c*{8}{c}}
\toprule
FFHQ & & ITER=0 & 100 & 500 & 1000 & 5000 & 10000 \\
\midrule
\multirow{4}{*}{\parbox{5cm}{\small\centering GaussianImage}} 
& PSNR & - & $10.4\pm1.7$ & $21.8\pm0.9$ & $29.4\pm1.6$ & $39.2\pm1.9$ & $40.1\pm1.9$ \\
& SSIM & - & $0.41\pm0.05$ & $0.77\pm0.05$ & $0.94\pm0.03$ & $0.99\pm0.001$ & $0.99\pm0.001$ \\
& Time (s) & - & $0.09\pm0.06$ & $0.45\pm0.06$ & $0.88\pm0.07$ & $6.36\pm0.08$ & $12.6\pm0.09$ \\
& \% & - & $0$ & $0$ & $0$ & $98$ & $99$ \\
\midrule
\multirow{4}{*}{\parbox{5cm}{\small\centering Ours (EigenGS with 300 components)}} 
& PSNR & $28.0$ & $34.4\pm2.4$ & $36.4\pm2.6$ & $37.5\pm2.6$ & $40.7\pm2.5$ & $41.8\pm2.4$ \\
& SSIM & $0.87$ & $0.95\pm0.02$ & $0.98\pm0.01$ & $0.99\pm0.01$ & $0.99\pm0.003$ & $0.99\pm0.002$\\
& Time (s) & - & $0.11\pm0.01$ & $0.53\pm0.02$ & $1.31\pm0.05$ & $5.92\pm0.03$ & $12.9\pm0.05$\\
& \% & - & $41$ & $76$ & $83$ & $98$ & $99$ \\
\midrule
\multirow{4}{*}{\parbox{5cm}{\small\centering Ours (EigenGS with 500 components)}} 
& PSNR & $28.9$ & $34.8\pm2.3$ & $36.7\pm2.6$ & $37.7\pm2.6$ & $40.8\pm2.5$ & $41.8\pm2.2$ \\
& SSIM & $0.88$ & $0.96\pm0.01$ & $0.98\pm0.01$ & $0.99\pm0.01$ & $0.99\pm0.004$ & $0.99\pm0.003$\\
& Time (s) & - & $0.25\pm0.02$ & $0.57\pm0.04$ & $1.14\pm0.02$ & $6.54\pm0.07$ & $12.8\pm0.11$\\
& \% & - & $50$ & $78$ & $84$ & $99$ & $99$ \\
\bottomrule
\end{tabular}
}
\caption{Quantitative comparison on the FFHQ dataset with 20{,}000 Gaussian points. 
The \textbf{\%} row indicates the percentage of test samples achieving PSNR larger than \textbf{35dB}.
Results along with the initialization scores of both 300 and 500 components EigenGS are included, presented with the mean and a standard deviation.
It is worth noting that the initial PSNR at iteration zero is roughly the same as the PSNR achieved by standard PCA with the same setting of the number of components as in our EigenGS approach.}
\label{tab:ffhq_result}
\end{table*}

\begin{figure*}
    \centering

    \begin{subfigure}[b]{0.23\textwidth}
        \includegraphics[width=\textwidth]{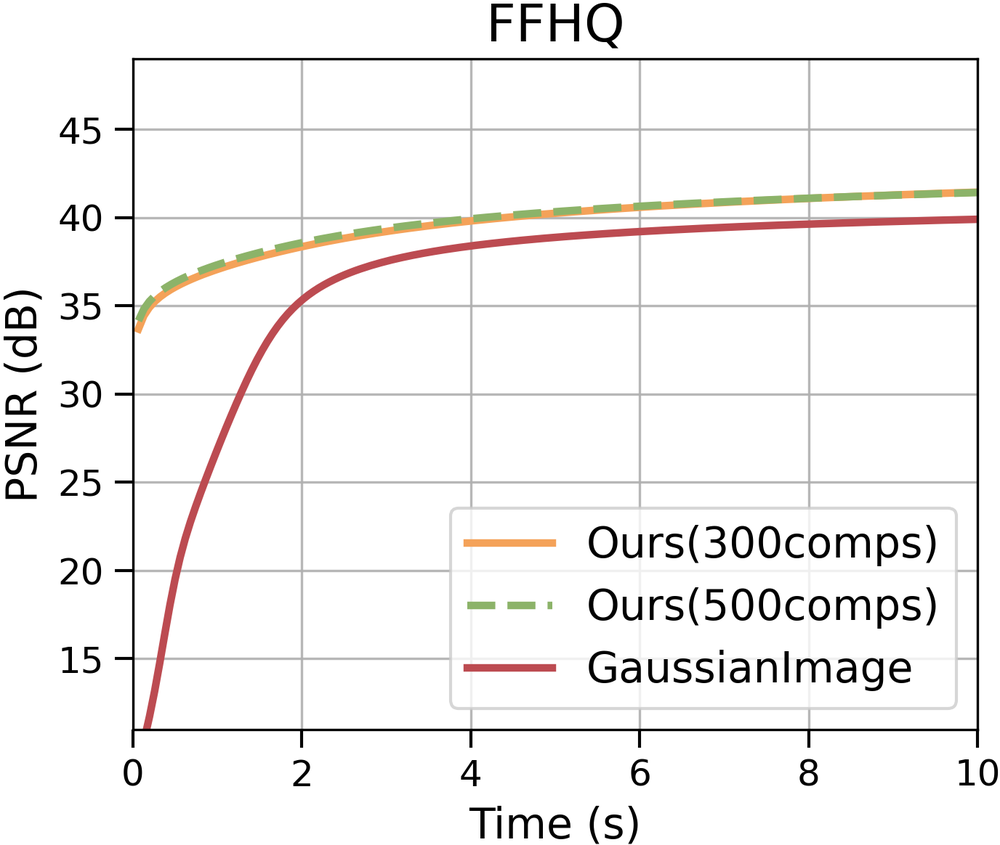}
    \end{subfigure}
    \begin{subfigure}[b]{0.23\textwidth}
        \includegraphics[width=\textwidth]{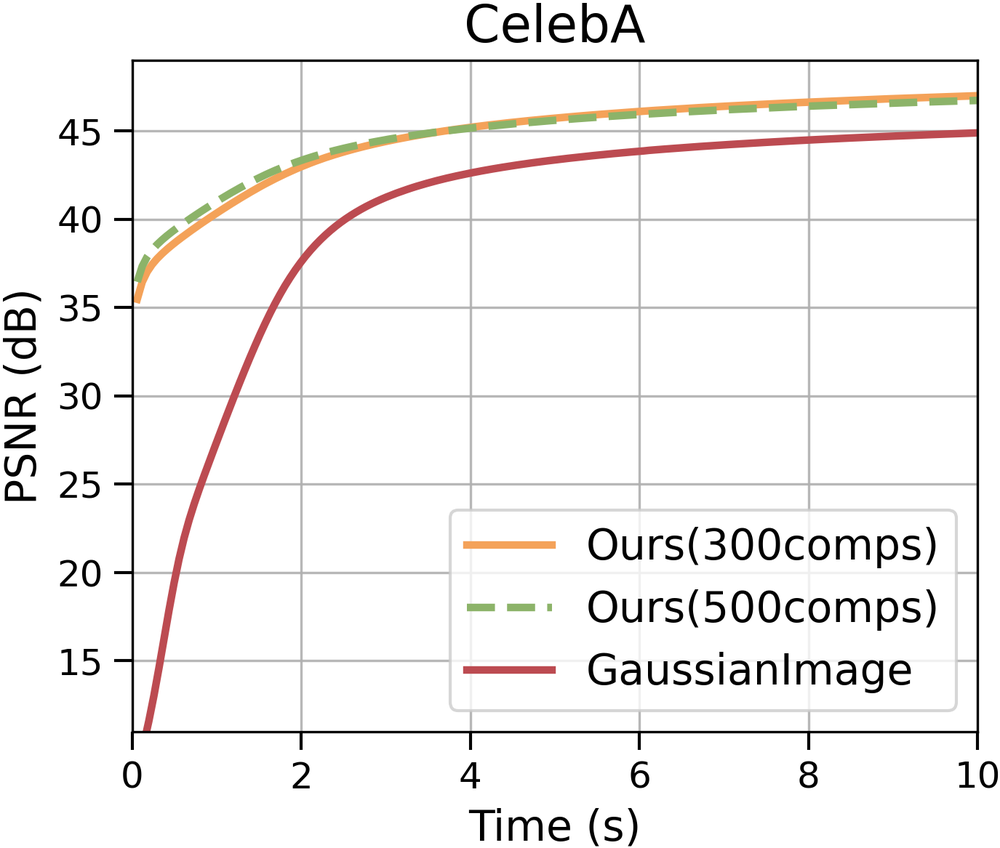}
    \end{subfigure}
    \begin{subfigure}[b]{0.23\textwidth}
        \includegraphics[width=\textwidth]{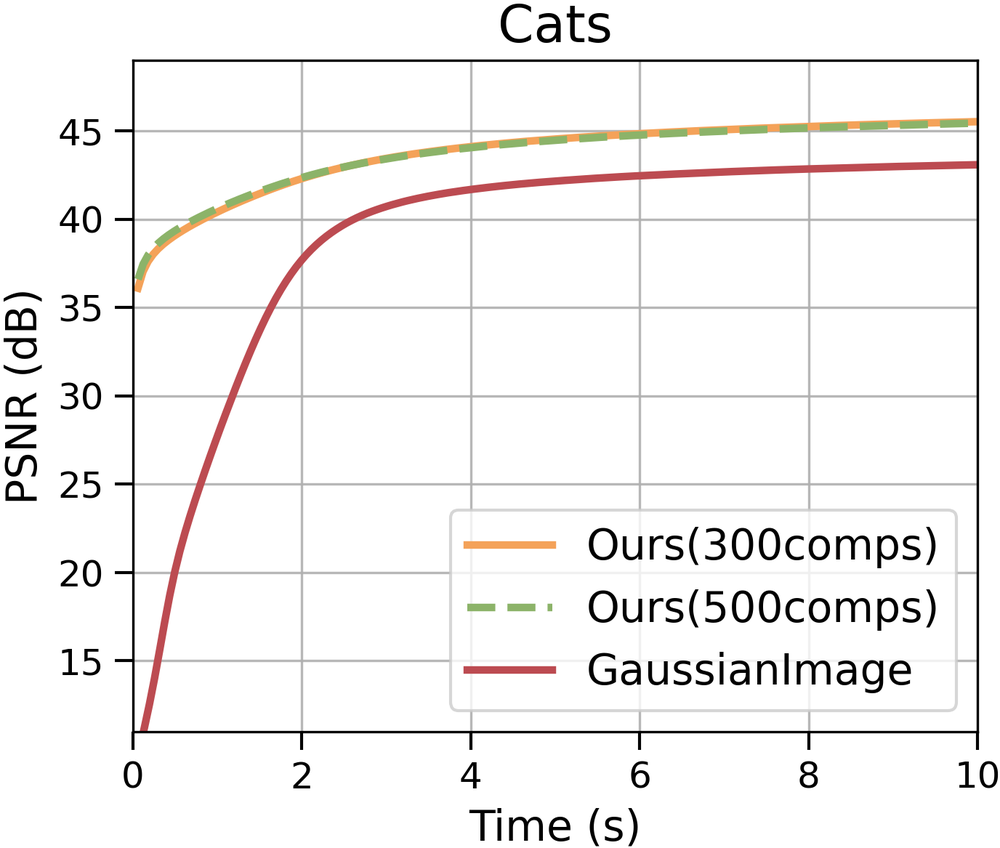}
    \end{subfigure}
    \begin{subfigure}[b]{0.23\textwidth}
        \includegraphics[width=\textwidth]{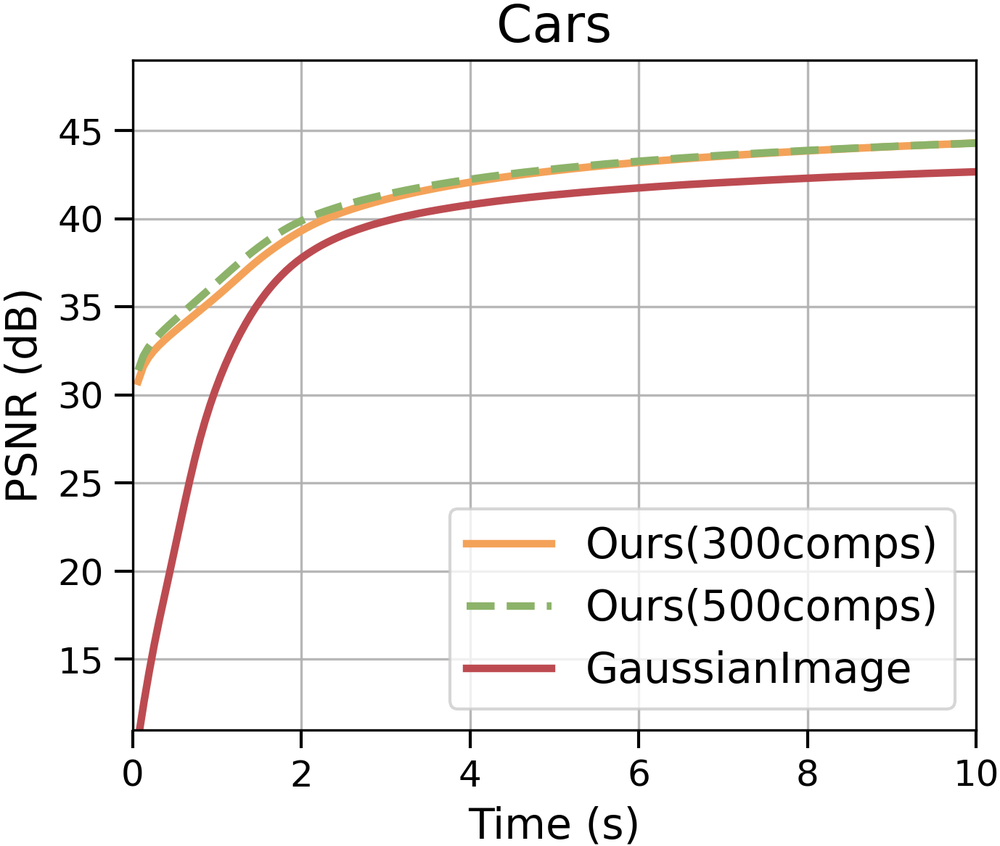}
    \end{subfigure}
    
    \caption{We sample the scores every 50 iterations on all datasets to exhibit the fast convergence capability, with particularly strong performance on smaller datasets like CelebA and Cats. The advantage is less pronounced but still evident on the more challenging Stanford Cars dataset due to its varied viewpoints and larger dimensions.}
    \label{fig:psnr_curve}
\end{figure*}

\begin{table*}
\centering
\resizebox{0.95\linewidth}{!}{%
\setlength{\tabcolsep}{2pt}
\begin{tabular}{c*{8}{c}}
\toprule
Cats \& CelebA & & ITER=0 & 100 & 500 & 1000 & 5000 & 10000 \\
\midrule
\multirow{3}{*}{\parbox{4.8cm}{\small\centering CelebA + GaussianImage}} 
& PSNR & - & $10.3\pm2.4$ & $21.9\pm1.2$ & $30.0\pm2.1$ & $43.8\pm2.9$ & $45.2\pm2.8$ \\
& SSIM & - & $0.47\pm0.05$ & $0.82\pm0.04$ & $0.95\pm0.02$ & $0.99\pm0.001$ & $0.99\pm0.001$ \\
& \% & - & $0$ & $0$ & $0$ & $91$ & $97$ \\
\midrule
\multirow{3}{*}{\parbox{4.8cm}{\small\centering CelebA + Cats-trained EigenGS}} 
& PSNR & $28.2$ & $36.2\pm3.7$ & $40.7\pm3.8$ & $43.2\pm3.5$ & $47.2\pm3.1$ & $48.2\pm3.1$ \\
& SSIM & $0.89$ & $0.96\pm0.03$ & $0.99\pm0.01$ & $0.99\pm0.003$ & $0.99\pm0.001$ & $0.99\pm0.001$\\
& \% & - & $17$ & $61$ & $84$ & $99$ & $100$ \\
\midrule
\multirow{3}{*}{\parbox{4.8cm}{\small\centering CelebA + ImageNet-trained EigenGS}} 
& PSNR & $28.7$ & $35.4\pm3.4$ & $39.6\pm3.7$ & $42.2\pm3.6$ & $46.3\pm3.2$ & $47.3\pm3.1$ \\
& SSIM & $0.91$ & $0.96\pm0.03$ & $0.99\pm0.01$ & $0.99\pm0.01$ & $0.99\pm0.002$ & $0.99\pm0.001$\\
& \% & - & $10$ & $47$ & $80$ & $98$ & $98$ \\
\midrule
\multirow{3}{*}{\parbox{4.8cm}{\small\centering Cats + ImageNet-trained EigenGS}} 
& PSNR & $29.6$ & $37.6\pm5.1$ & $41.1\pm5.6$ & $42.9\pm5.4$ & $45.9\pm4.7$ & $46.6\pm4.6$ \\
& SSIM & $0.89$ & $0.96\pm0.04$ & $0.99\pm0.01$ & $0.99\pm0.01$ & $0.99\pm0.002$ & $0.99\pm0.001$\\
& \% & - & $35$ & $59$ & $67$ & $88$ & $90$ \\
\bottomrule
\end{tabular}
}
\caption{Cross-dataset experiments showing PSNR, SSIM, and percentage of images achieving PSNR larger than \textbf{40dB} at different iterations. We evaluate two scenarios: applying CelebA-trained EigenGS to the Cats dataset and vice versa. The results still show advantages, especially within 1000 iterations.}
\vspace{0.15cm}
\label{tab:transfer_result}
\end{table*}

\subsection{Datasets and Metrics}

We conduct extensive experiments on several widely used datasets that cover various object categories while maintaining instance diversity. Our primary experiments are carried out on face datasets at different resolutions: CelebA~\cite{LiuLWT15} with 256×256 pixel images and FFHQ~\cite{KarrasLA19} with higher resolution 512×512 pixel images. To assess generalization and provide a more comprehensive understanding, we include datasets with other object types, such as the Cats Dataset~\cite{ZhangST08} and the Stanford Cars Dataset~\cite{Krause0DF13}. 

A training set comprising 10,000 images is constructed for each dataset to facilitate the PCA decomposition. Model performance pertaining to each dataset is then evaluated on a separate test set of 100 randomly selected images for quantitative evaluation. This experimental setup ensures rigorous quantitative assessments while maintaining computational feasibility for comprehensive comparisons.

We first report the detailed result of FFHQ in \Cref{tab:ffhq_result}. The \% at the last row for each setup denotes the percentage of test samples successfully achieving the targeted PSNR threshold. For the other three datasets, due to space constraints, we only plot the PSNR curves as shown in \Cref{fig:psnr_curve} with the sampling rate of every 50 iterations. This analysis particularly highlights the accelerated convergence attained by our method compared to GaussianImage~\cite{ZhangGXHWQLGZ24}.

For a fair comparison with the baseline method, GaussianImage, we maintain consistent experimental settings using the same number of Gaussians, which is 5000 points on a smaller dataset and 20,000 points on a larger one, and the same 10,000 iteration counts across all tests. All experiments are conducted on a single V100 GPU.

\subsection{Quantitative Results}

Our comprehensive experimental evaluation demonstrates the significant advantages of EigenGS across multiple dimensions. The results on the FFHQ dataset, as shown in \Cref{tab:ffhq_result}, demonstrate superior performance in both early-stage convergence and final reconstruction quality. The effectiveness of our PCA-based initialization strategy is immediately apparent, with starting PSNR values of 28.0 dB and 28.9 dB for the 300- and 500-component configurations, respectively. This represents a dramatic improvement over the random initialization employed by GaussianImage, which essentially starts from noise. This strong initialization is crucial for practical applications of Gaussian Splatting to images, as it enables meaningful visual results from the very beginning of the optimization process.

The advantage of our initialization method is especially prominent at the early stage of training. For both configurations of 300 and 500 components, our approach demonstrates remarkable efficiency in that over 80\% of the samples reach a threshold PSNR of 35 dB within 1000 iterations. In contrast, GaussianImage requires substantially more iterations to reach comparable quality levels, highlighting the advantages of our approach. Our method ultimately delivers a higher final PSNR, with the 300-component configuration achieving a PSNR of 41.8 dB and the 500-component configuration reaching a final PSNR of 41.7 dB. These results not only surpass GaussianImage but also demonstrate an interesting insight: the number of components primarily affects early-stage convergence rather than final quality. This suggests that our method efficiently captures the essential image characteristics regardless of the initial component count.

The PSNR curves in \Cref{fig:psnr_curve} further validate our method's effectiveness across diverse datasets with varying characteristics. The well-aligned nature of CelebA and the consistent structure of cat faces, along with their moderate resolution, allow our PCA-based initialization to provide particularly effective starting points. Although the Stanford Cars dataset presents a more challenging scenario due to its larger image dimensions and less aligned nature, with vehicles photographed from various angles and distances, our method still demonstrates clear advantages, albeit with a smaller margin compared to the face datasets. Across all datasets, our approach maintains consistent superiority in both convergence speed and final reconstruction quality, as evidenced by the steeper initial slopes in the PSNR curves compared to GaussianImage.

\begin{table*}
\centering

\begin{tabular}{lcccccccc}
\toprule
& \multicolumn{2}{c}{CelebA} 
& \multicolumn{2}{c}{FFHQ} 
& \multicolumn{2}{c}{Cats} 
& \multicolumn{2}{c}{Cars} \\
\cmidrule(lr){2-3} \cmidrule(lr){4-5} \cmidrule(lr){6-7} \cmidrule(lr){8-9}
\parbox{3.5cm}{\centering Ablation Setup} 
& PSNR$\uparrow$ & SSIM$\uparrow$
& PSNR$\uparrow$ & SSIM$\uparrow$
& PSNR$\uparrow$ & SSIM$\uparrow$
& PSNR$\uparrow$ & SSIM$\uparrow$ \\
\midrule
\multirow{1}{*}{\parbox{3.5cm}{Ours-YCbCr}} 
& 47.2 & 0.998
& \textbf{41.8} & \textbf{0.994}
& 45.7 & 0.997
& \textbf{44.7} & \textbf{0.998} \\
\multirow{1}{*}{\parbox{3.5cm}{Ours-YCbCr (w/o FL)}} 
& \textbf{48.0} & \textbf{0.998}
& 40.7 & 0.992
& \textbf{46.1} & \textbf{0.997}
& 43.5 & 0.996 \\
\multirow{1}{*}{\parbox{3.5cm}{Ours-RGB}} 
& 39.5 & 0.994
& 34.9 & 0.986
& 38.5 & 0.992
& 36.4 & 0.994 \\
\multirow{1}{*}{\parbox{3.5cm}{Ours-RGB (w/o FL)}} 
& 39.9 & 0.995
& 33.3 & 0.976
& 38.9 & 0.993
& 35.1 & 0.992 \\
\bottomrule
\end{tabular}
\caption{Ablation studies on our method across different datasets comparing YCbCr and RGB color spaces, with and without frequency-aware learning (FL). Reported values show final PSNR (dB) and SSIM with total 10,000 iterations of training. }
\vspace{-0.2cm}
\label{tab:ablation}
\end{table*}

\subsection{Cross-Dataset Generalization}
\begin{figure}[t]
    \centering

    \begin{subfigure}[b]{0.15\textwidth}
        \includegraphics[width=\textwidth]{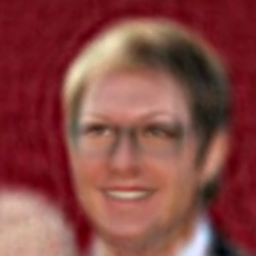}
    \end{subfigure}
    \begin{subfigure}[b]{0.15\textwidth}
        \includegraphics[width=\textwidth]{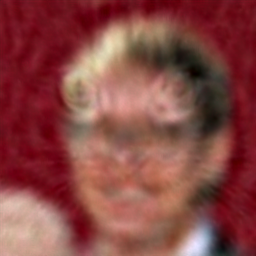}
    \end{subfigure}
    \begin{subfigure}[b]{0.15\textwidth}
        \includegraphics[width=\textwidth]{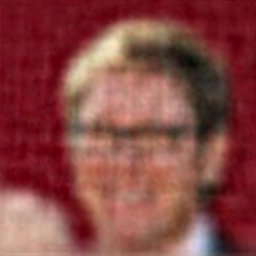}
    \end{subfigure}
    
    \begin{subfigure}[b]{0.15\textwidth}
        \includegraphics[width=\textwidth]{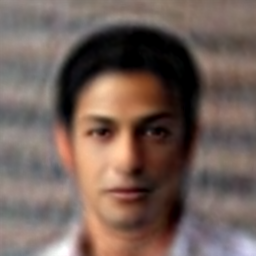}
    \end{subfigure}
    \begin{subfigure}[b]{0.15\textwidth}
        \includegraphics[width=\textwidth]{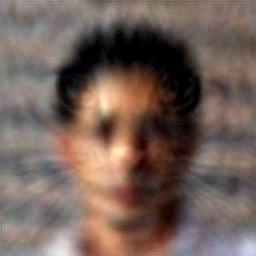}
    \end{subfigure}
    \begin{subfigure}[b]{0.15\textwidth}
        \includegraphics[width=\textwidth]{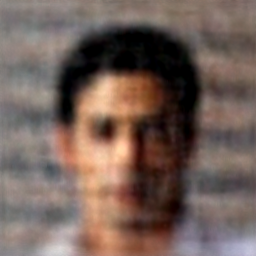}
    \end{subfigure}

    \caption{Cross-dataset initialization comparison. Left Column: initialization using EigenGS from the corresponding dataset. Middle column: initialization using Cats-trained EigenGS for CelebA dataset. Right Column: initialization using ImageNet-trained EigenGS for CelebA dataset. Despite some domain-specific artifacts in cross-dataset results, it still demonstrates effective color and structure preservation. Therefore, our method can still achieve high-quality reconstruction even with cross-dataset initialization, which aligns with the quantitative evaluations shown in \Cref{tab:transfer_result}.}
    \label{fig:cross}
\end{figure}

To assess the capabilities of \emph{transfer generalization} of EigenGS, we perform cross-dataset experiments between the CelebA and Cats datasets. Specifically, we examine two scenarios: using EigenGS trained on the Cats dataset to initialize representations for CelebA images and, conversely, using CelebA-trained EigenGS to initialize Cats images. The results, presented in \Cref{tab:transfer_result}, demonstrate the robust transfer capabilities of our method. Remarkably, even when applying eigen-Gaussians across distinctly different image categories, EigenGS maintains superior performance compared to GaussianImage's random initialization.

A visual inspection of the EigenGS initializations reveals an interesting phenomenon. While the initial reconstructions may exhibit faint ``shadow'' artifacts from the source domain, such as cat-like features in facial reconstructions or facial features in cat images, the preservation of color and overall structure remains strong. This suggests that the EigenGS representations can capture useful universal image statistics that transfer effectively across datasets, particularly in terms of color distribution and fundamental structural elements. These results demonstrate the robust transfer capabilities across distinct image domains without specialized retraining or domain adaptation techniques.

To further validate our observation, we train EigenGS on a diverse set of images from ImageNet \cite{DengDSLL009}, which presents a significantly more challenging scenario with its ``in-the-wild'' characteristics, including unaligned images of varying sizes, aspect ratios, and content categories. Using the exactly same experimental settings as in our previous tests, the results \Cref{tab:transfer_result} are particularly impressive, which again demonstrates that our method can effectively break free from traditional PCA limitations, learning a basis that captures the global luminance and color features, regardless of the unstructured characteristics.

The success of the ImageNet-trained model is especially noteworthy because it suggests that given a training set with sufficient color and structural diversity, we can develop a more universal set of eigen-Gaussians that effectively captures generalizable representations of both image structure and color. This capability becomes particularly valuable when combined with the Gaussian pipeline's optimization capabilities, which can further refine the initial representation to achieve high-quality results. 

Results of cross-domain experiments particularly reinforce our hypothesis that with comprehensive training data and an appropriate number of components, we can potentially develop a general-purpose set of EigenGS that transcends category-specific limitations, serving as an effective initialization technique for Gaussian-based methods

\subsection{Ablation Analysis}

We conduct ablation studies on the 300-component configuration for all datasets to analyze two key design choices in our method. The first investigates the impact of color space selection, comparing conventional RGB representation against YCbCr space, where luminance and chrominance information are processed independently. The second examines our frequency-aware learning (FL) scheme, which strategically allocates a portion of Gaussians to model low-frequency components, enabling specialized representation across different frequency bands. \Cref{tab:ablation} summarizes these experimental results, demonstrating the significance of each component in our pipeline.

Our experiments demonstrate consistent performance advantages when using YCbCr color space over RGB across all datasets. This improvement can be attributed to two key factors. First, YCbCr's separation of luminance (Y: 16-235) and chrominance (Cb/Cr: 16-240) components provide natural headroom for handling out-of-bounds values that arise from PCA-based reconstruction, particularly for test set images. Second, since color channels are inherently more susceptible to outliers than intensity channels, YCbCr's reduced number of color channels makes it naturally more robust when combined with PCA-based methods. Thus, the choice of YCbCr color space helps minimize performance degradation from value truncation, as reflected in the significantly improved PSNR results.

Compared to the consistent advantages of YCbCr color space across all datasets, the impact of frequency learning exhibits more complex patterns. The effectiveness of FL indicated a notable correlation with image resolution and content characteristics. It provides moderate but important advantages in datasets like FFHQ and Cars, which has higher quality and larger resolution.

Although these average improvements might appear less significant than those observed in the color space ablation, they address a critical issue: the penny-round-tile artifacts that can severely degrade image quality in certain scenarios. The impact of these artifacts is not uniform across all test images. They primarily show up in regions with smooth gradients, such as the textureless areas of lighter tones, where they can cause PSNR degradations of 3-4 dB on these specific test cases if left unaddressed. Thus, while the aggregate PSNR improvements may seem modest, FL plays an essential role in preventing quality degradation in susceptible images, making it a necessary component for robust, high-quality reconstruction.

\begin{figure}
    \centering

        \includegraphics[width=0.23\textwidth]{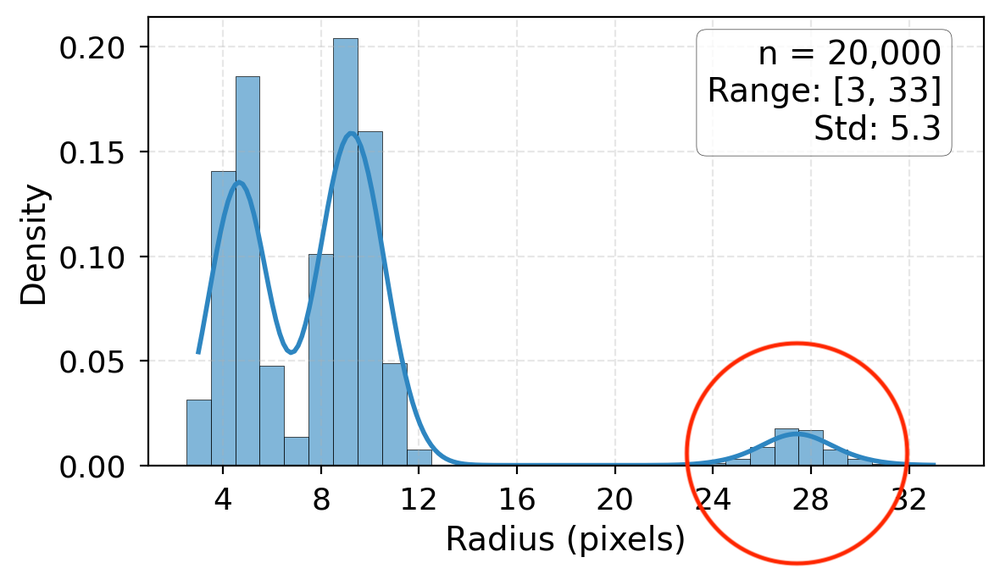}
        \includegraphics[width=0.23\textwidth]{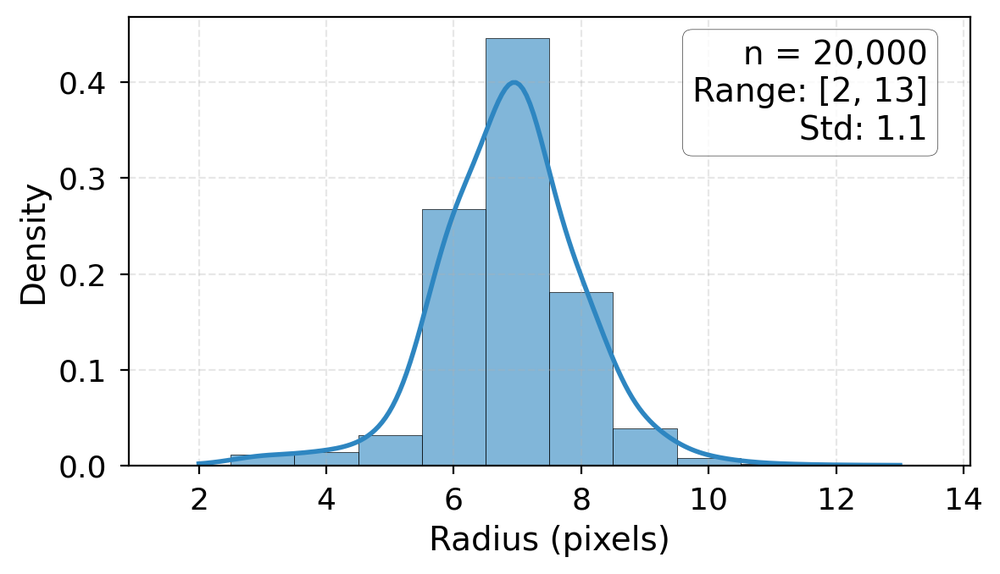}
    
    \caption{Distribution of Gaussian sizes shows the effectiveness of FL. The x-axis represents the Gaussian size in pixels, and the y-axis represents the density of Gaussians at each radius value. Without FL (right): Gaussians converge to uniformly small radius, resulting in limited size diversity. With FL (left): a small group of large Gaussians emerge for low-frequency content. This size diversity helps prevent ``penny-round-tile" artifacts and better represents different spatial frequencies in the image.}
    \vspace{-0.2cm}
    \label{fig:radii}
\end{figure}

\begin{figure*}
    \centering
    \begin{minipage}{0.19\textwidth}
        \centering
        ITER = 0
    \end{minipage}%
    \begin{minipage}{0.19\textwidth}
        \centering
        ITER = 10
    \end{minipage}%
    \begin{minipage}{0.19\textwidth}
        \centering
        ITER = 100
    \end{minipage}%
    \begin{minipage}{0.19\textwidth}
        \centering
        ITER = 1000
    \end{minipage}%
    \begin{minipage}{0.19\textwidth}
        \centering
        GT
    \end{minipage}
    \vspace{0.3em}

    \begin{subfigure}[b]{0.19\textwidth}
        \includegraphics[width=\textwidth]{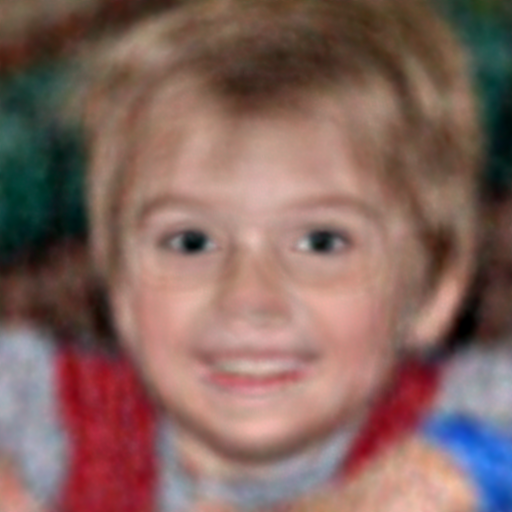}
    \end{subfigure}
    \begin{subfigure}[b]{0.19\textwidth}
        \includegraphics[width=\textwidth]{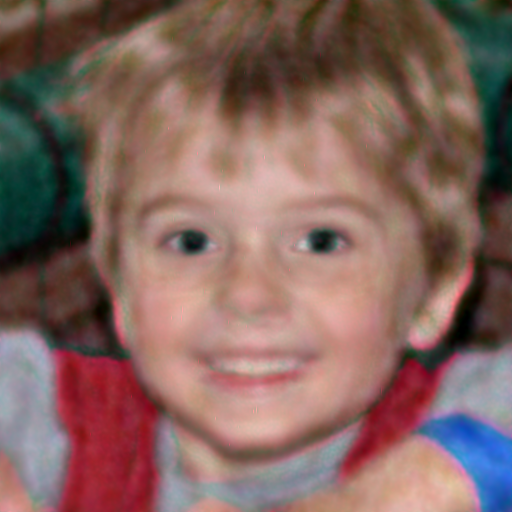}
    \end{subfigure}
    \begin{subfigure}[b]{0.19\textwidth}
        \includegraphics[width=\textwidth]{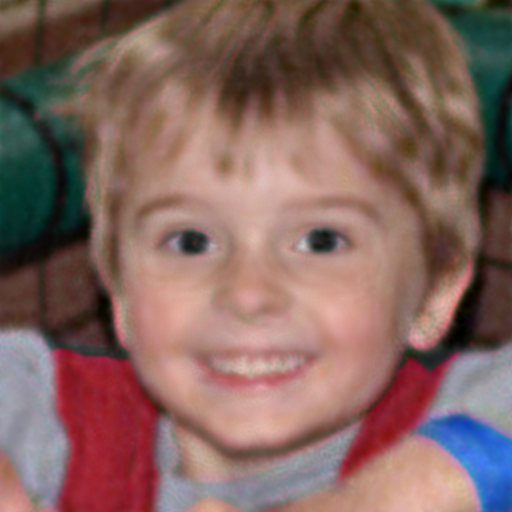}
    \end{subfigure}
    \begin{subfigure}[b]{0.19\textwidth}
        \includegraphics[width=\textwidth]{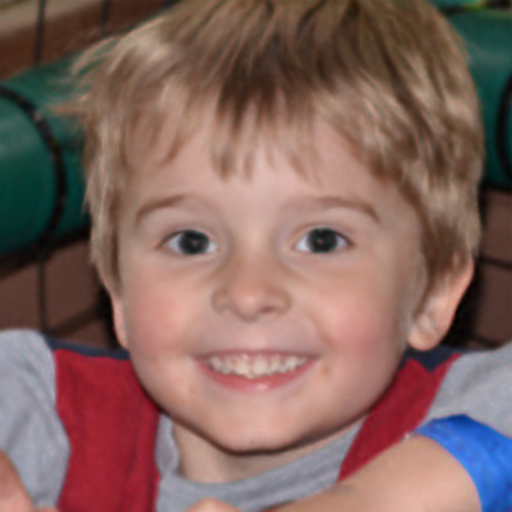}
    \end{subfigure}
    \begin{subfigure}[b]{0.19\textwidth}
        \includegraphics[width=\textwidth]{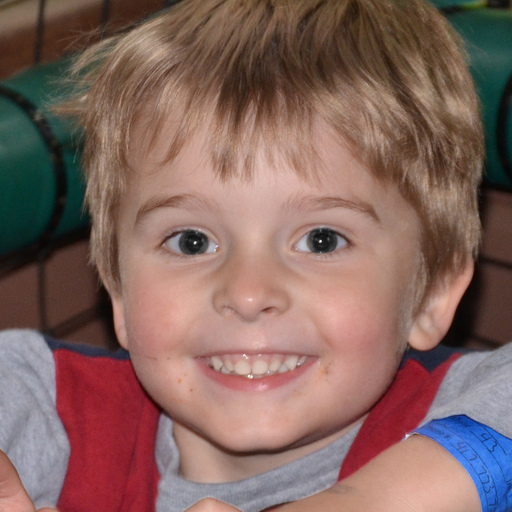}
    \end{subfigure}

    \begin{subfigure}[b]{0.19\textwidth}
        \includegraphics[width=\textwidth]{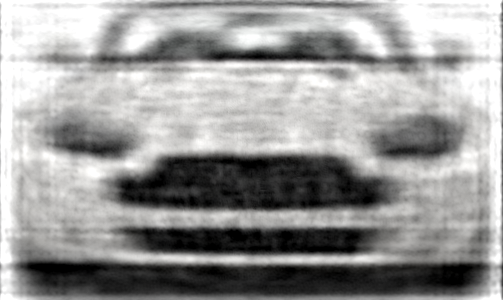}
    \end{subfigure}
    \begin{subfigure}[b]{0.19\textwidth}
        \includegraphics[width=\textwidth]{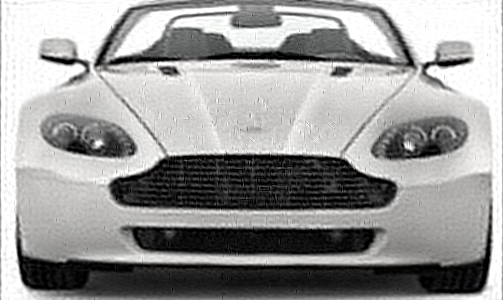}
    \end{subfigure}
    \begin{subfigure}[b]{0.19\textwidth}
        \includegraphics[width=\textwidth]{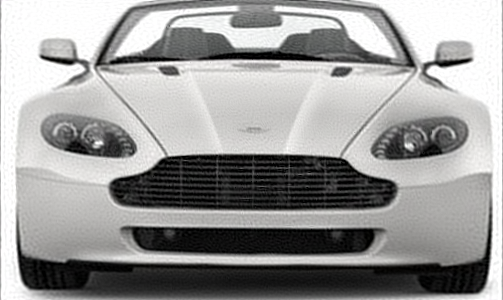}
    \end{subfigure}
    \begin{subfigure}[b]{0.19\textwidth}
        \includegraphics[width=\textwidth]{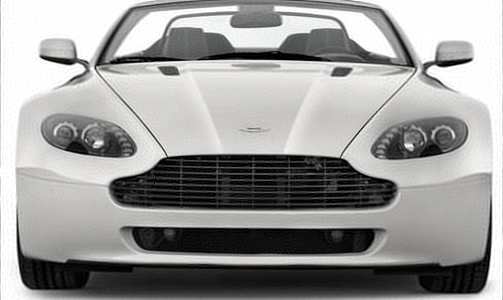}
    \end{subfigure}
    \begin{subfigure}[b]{0.19\textwidth}
        \includegraphics[width=\textwidth]{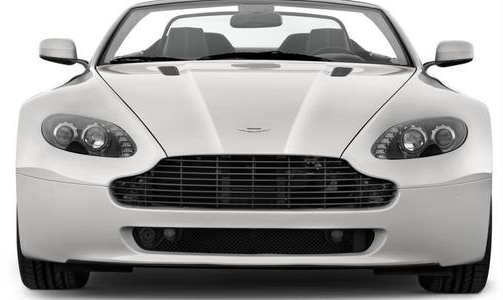}
    \end{subfigure}

    \begin{subfigure}[b]{0.19\textwidth}
        \includegraphics[width=\textwidth]{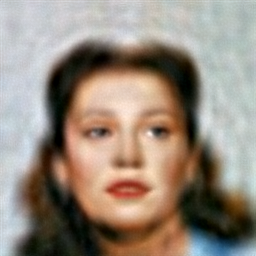}
    \end{subfigure}
    \begin{subfigure}[b]{0.19\textwidth}
        \includegraphics[width=\textwidth]{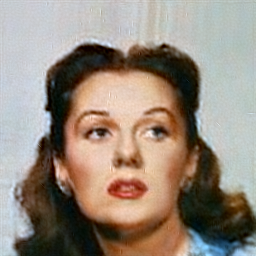}
    \end{subfigure}
    \begin{subfigure}[b]{0.19\textwidth}
        \includegraphics[width=\textwidth]{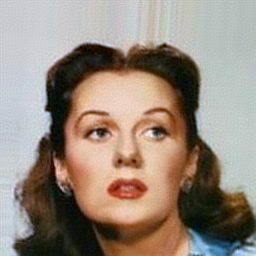}
    \end{subfigure}
    \begin{subfigure}[b]{0.19\textwidth}
        \includegraphics[width=\textwidth]{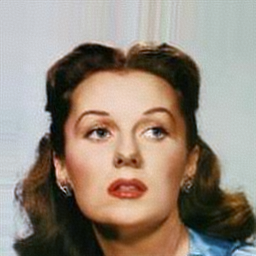}
    \end{subfigure}
    \begin{subfigure}[b]{0.19\textwidth}
        \includegraphics[width=\textwidth]{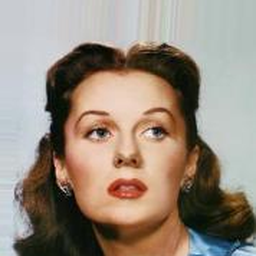}
    \end{subfigure}

    \begin{subfigure}[b]{0.19\textwidth}
        \includegraphics[width=\textwidth]{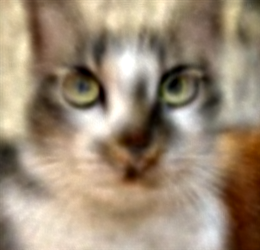}
    \end{subfigure}
    \begin{subfigure}[b]{0.19\textwidth}
        \includegraphics[width=\textwidth]{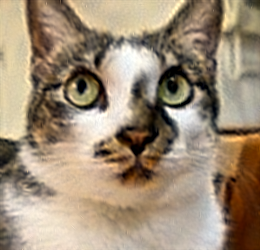}
    \end{subfigure}
    \begin{subfigure}[b]{0.19\textwidth}
        \includegraphics[width=\textwidth]{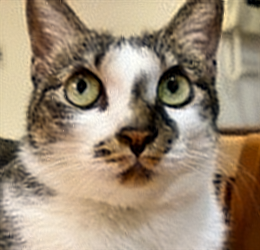}
    \end{subfigure}
    \begin{subfigure}[b]{0.19\textwidth}
        \includegraphics[width=\textwidth]{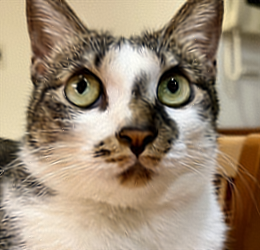}
    \end{subfigure}
    \begin{subfigure}[b]{0.19\textwidth}
        \includegraphics[width=\textwidth]{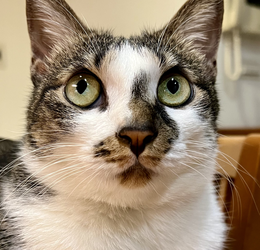}
    \end{subfigure}

    \caption{Qualitative results from various datasets. From top to bottom: FFHQ (PSNR: 25, 30, 33, 38), Stanford Cars (PSNR: 30, 32, 34, 39), CelebA (PSNR: 30, 36, 40, 45), and Cats datasets (PSNR: 25, 29, 31, 34), respectively. Overall, we can get quality results within 1000 iterations in most cases, though different convergence rates for each test sample can be observed here.}
    \vspace{-0.1cm}
    \label{fig:comparison}
\end{figure*}

Nevertheless, our experiments reveal that the impact of FL is not uniformly positive across all scenarios. In smaller resolution datasets such as CelebA and the Cats dataset, we observe slight performance decreases of -0.8 dB and -0.4 dB, respectively. \Cref{fig:radii} shows that FL enforces a bimodal distribution of Gaussian sizes, with a dedicated portion of Gaussians maintaining larger sizes for low-frequency content. This intended allocation, while essential for preventing artifacts in high-resolution images, comes with a trade-off: fewer Gaussians are available for general reconstruction purposes compared to the version without FL where all Gaussians can freely converge to uniformly small sizes. 

In lower-resolution images where penny-round-tile artifacts are less prominent, the reduction in flexibility of Gaussian allocation can lead to slightly decreased reconstruction quality. These findings suggest the application of FL should be considered based on dataset characteristics, particularly image resolution and content type, to balance the prevention of artifacts and overall reconstruction fidelity.

\section{Conclusion}

We introduce EigenGS, a novel approach that successfully bridges classical dimensionality reduction techniques with modern Gaussian Splatting methods. The primary contributions of our work are threefold. First, we resolve the initialization challenge in Gaussian Splatting by leveraging PCA-derived EigenGS, facilitating instant and effective initializations for modeling new images. Second, the proposed frequency-aware learning mechanism effectively prevents the "penny-round-tile" by encouraging the learned Gaussians to adapt to different spatial frequencies. Third, the success of cross-dataset experiments, especially with ImageNet-trained models, reveals the potential for developing general-purpose representations that transcend specific image categories.

This work demonstrates that classical computer vision techniques can be effectively combined with modern approaches to create more efficient and robust solutions. The synergy between PCA's efficient dimensionality reduction and Gaussian Splatting's high-quality reconstruction capabilities suggests that established methods merit renewed consideration as novel techniques are introduced.

{
    \small
    \bibliographystyle{ieeenat_fullname}
    \bibliography{main}
}

\clearpage
\renewcommand\thepage{\roman{page}}
\setcounter{page}{1}
\maketitlesupplementary
\renewcommand\thesection{\Roman{section}}
\setcounter{section}{0}
\renewcommand\thesubsection{\roman{subsection}}
\setcounter{subsection}{0}
\renewcommand\thefigure{\Roman{figure}}
\setcounter{figure}{0}
\renewcommand\thetable{\Roman{table}}
\setcounter{table}{0}

\section*{Additional Results \& Visualizations}
The supplementary material includes additional visualizations and detailed experimental results that highlight key aspects of our method, where the visualization is primarily using the FFHQ dataset. All experimental settings are aligned with those described in the main paper. \Cref{sec:6.3} presents comprehensive quantitative and qualitative results across different datasets. Our method is able to achieve consistent performance across diverse image categories and resolutions.
\Cref{sec:6.1} provides an analysis of our frequency-aware learning mechanism. We show the advantage of color space selection in \Cref{sec:6.2}.

\section{Quantitative and Qualitative Evaluations}
\label{sec:6.3}

\begin{figure*}
    \centering
    \begin{minipage}{0.19\textwidth}
        \centering
        ITER = 0
    \end{minipage}%
    \begin{minipage}{0.19\textwidth}
        \centering
        ITER = 10
    \end{minipage}%
    \begin{minipage}{0.19\textwidth}
        \centering
        ITER = 100
    \end{minipage}%
    \begin{minipage}{0.19\textwidth}
        \centering
        ITER = 1000
    \end{minipage}%
    \begin{minipage}{0.19\textwidth}
        \centering
        GT
    \end{minipage}
    \vspace{0.3em}

    \begin{subfigure}[b]{0.19\textwidth}
        \includegraphics[width=\textwidth]{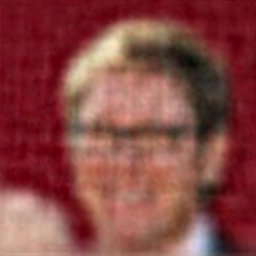}
    \end{subfigure}
    \begin{subfigure}[b]{0.19\textwidth}
        \includegraphics[width=\textwidth]{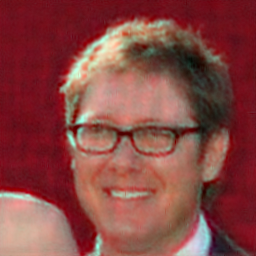}
    \end{subfigure}
    \begin{subfigure}[b]{0.19\textwidth}
        \includegraphics[width=\textwidth]{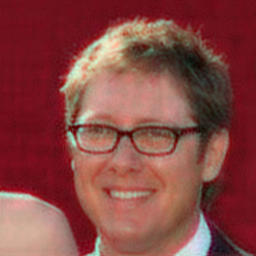}
    \end{subfigure}
    \begin{subfigure}[b]{0.19\textwidth}
        \includegraphics[width=\textwidth]{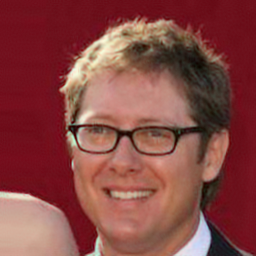}
    \end{subfigure}
    \begin{subfigure}[b]{0.19\textwidth}
        \includegraphics[width=\textwidth]{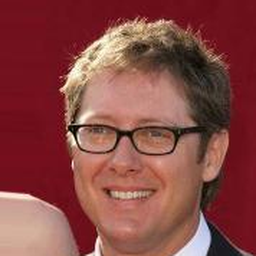}
    \end{subfigure}

    \begin{subfigure}[b]{0.19\textwidth}
        \includegraphics[width=\textwidth]{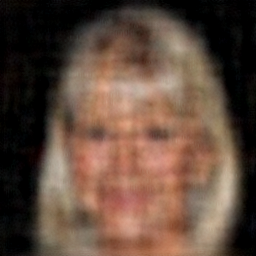}
    \end{subfigure}
    \begin{subfigure}[b]{0.19\textwidth}
        \includegraphics[width=\textwidth]{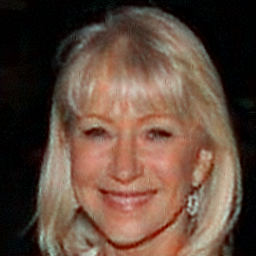}
    \end{subfigure}
    \begin{subfigure}[b]{0.19\textwidth}
        \includegraphics[width=\textwidth]{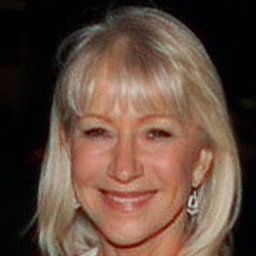}
    \end{subfigure}
    \begin{subfigure}[b]{0.19\textwidth}
        \includegraphics[width=\textwidth]{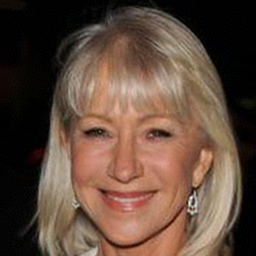}
    \end{subfigure}
    \begin{subfigure}[b]{0.19\textwidth}
        \includegraphics[width=\textwidth]{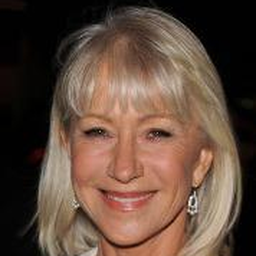}
    \end{subfigure}

    \begin{subfigure}[b]{0.19\textwidth}
        \includegraphics[width=\textwidth]{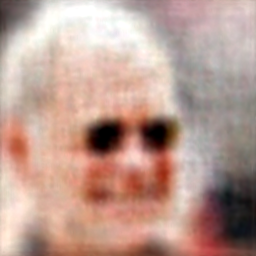}
    \end{subfigure}
    \begin{subfigure}[b]{0.19\textwidth}
        \includegraphics[width=\textwidth]{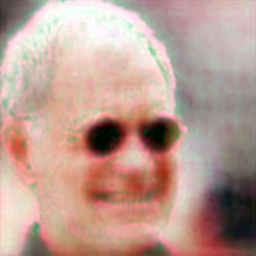}
    \end{subfigure}
    \begin{subfigure}[b]{0.19\textwidth}
        \includegraphics[width=\textwidth]{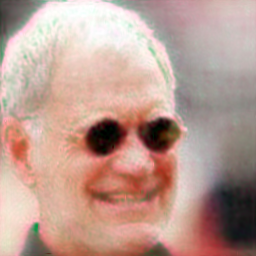}
    \end{subfigure}
    \begin{subfigure}[b]{0.19\textwidth}
        \includegraphics[width=\textwidth]{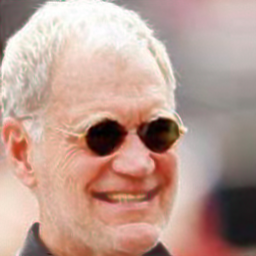}
    \end{subfigure}
    \begin{subfigure}[b]{0.19\textwidth}
        \includegraphics[width=\textwidth]{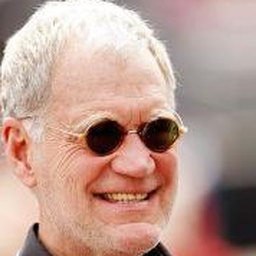}
    \end{subfigure}

    \begin{subfigure}[b]{0.19\textwidth}
        \includegraphics[width=\textwidth]{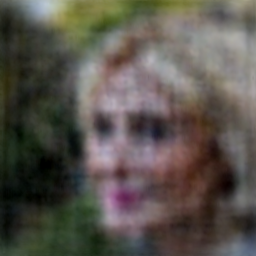}
    \end{subfigure}
    \begin{subfigure}[b]{0.19\textwidth}
        \includegraphics[width=\textwidth]{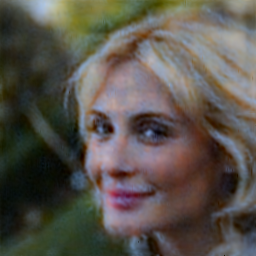}
    \end{subfigure}
    \begin{subfigure}[b]{0.19\textwidth}
        \includegraphics[width=\textwidth]{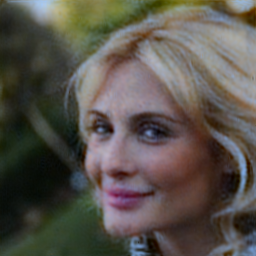}
    \end{subfigure}
    \begin{subfigure}[b]{0.19\textwidth}
        \includegraphics[width=\textwidth]{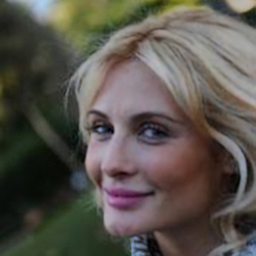}
    \end{subfigure}
    \begin{subfigure}[b]{0.19\textwidth}
        \includegraphics[width=\textwidth]{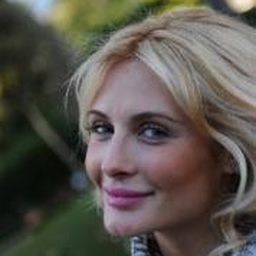}
    \end{subfigure}

    \begin{subfigure}[b]{0.19\textwidth}
        \includegraphics[width=\textwidth]{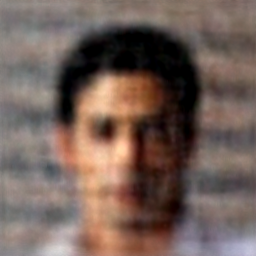}
    \end{subfigure}
    \begin{subfigure}[b]{0.19\textwidth}
        \includegraphics[width=\textwidth]{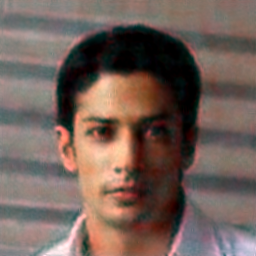}
    \end{subfigure}
    \begin{subfigure}[b]{0.19\textwidth}
        \includegraphics[width=\textwidth]{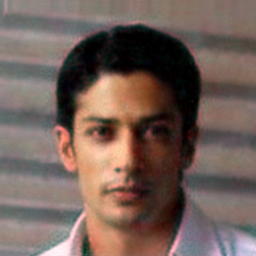}
    \end{subfigure}
    \begin{subfigure}[b]{0.19\textwidth}
        \includegraphics[width=\textwidth]{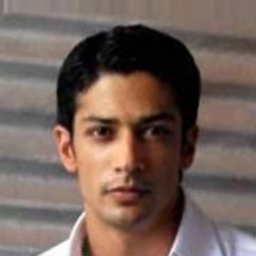}
    \end{subfigure}
    \begin{subfigure}[b]{0.19\textwidth}
        \includegraphics[width=\textwidth]{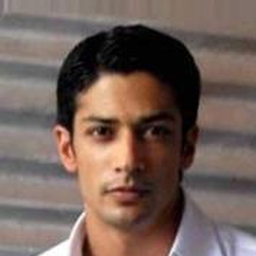}
    \end{subfigure}

    \caption{Qualitative results on the CelebA dataset using ImageNet-trained EigenGS.}
    \label{fig:suppl_celeba_vis}
\end{figure*}

\begin{figure*}
    \centering
    \begin{minipage}{0.19\textwidth}
        \centering
        ITER = 0
    \end{minipage}%
    \begin{minipage}{0.19\textwidth}
        \centering
        ITER = 10
    \end{minipage}%
    \begin{minipage}{0.19\textwidth}
        \centering
        ITER = 100
    \end{minipage}%
    \begin{minipage}{0.19\textwidth}
        \centering
        ITER = 1000
    \end{minipage}%
    \begin{minipage}{0.19\textwidth}
        \centering
        GT
    \end{minipage}
    \vspace{0.3em}

    \begin{subfigure}[b]{0.19\textwidth}
        \includegraphics[width=\textwidth]{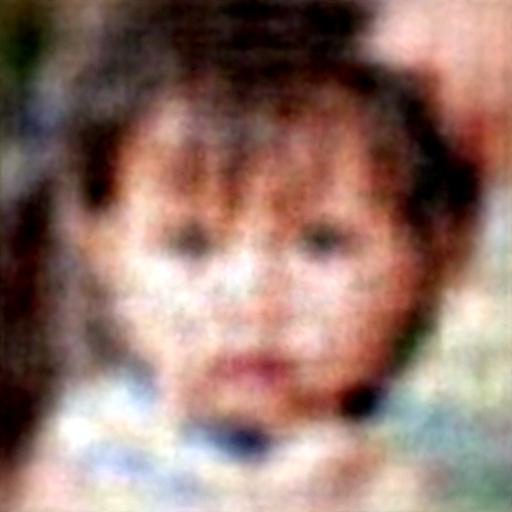}
    \end{subfigure}
    \begin{subfigure}[b]{0.19\textwidth}
        \includegraphics[width=\textwidth]{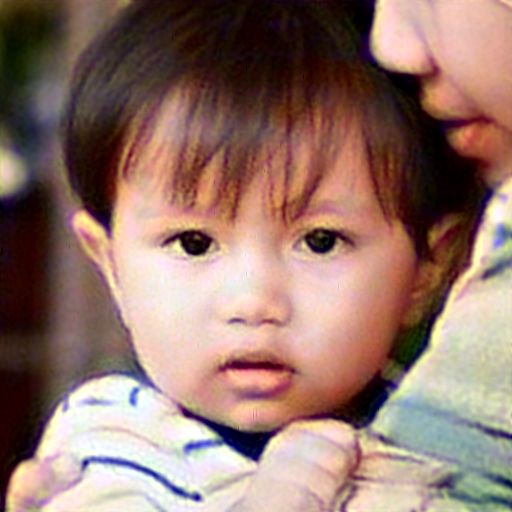}
    \end{subfigure}
    \begin{subfigure}[b]{0.19\textwidth}
        \includegraphics[width=\textwidth]{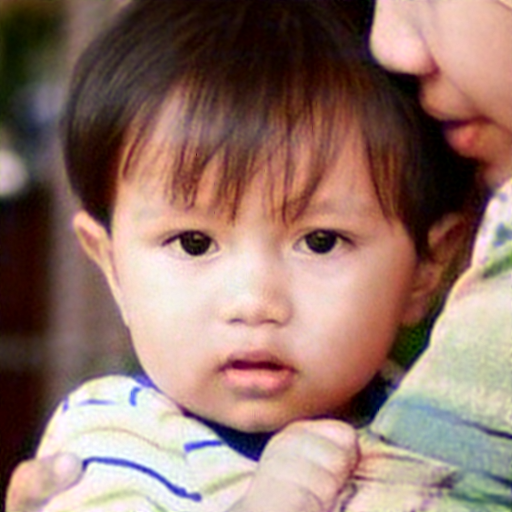}
    \end{subfigure}
    \begin{subfigure}[b]{0.19\textwidth}
        \includegraphics[width=\textwidth]{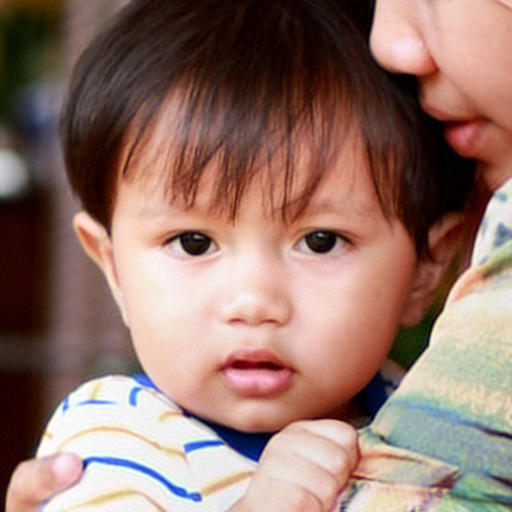}
    \end{subfigure}
    \begin{subfigure}[b]{0.19\textwidth}
        \includegraphics[width=\textwidth]{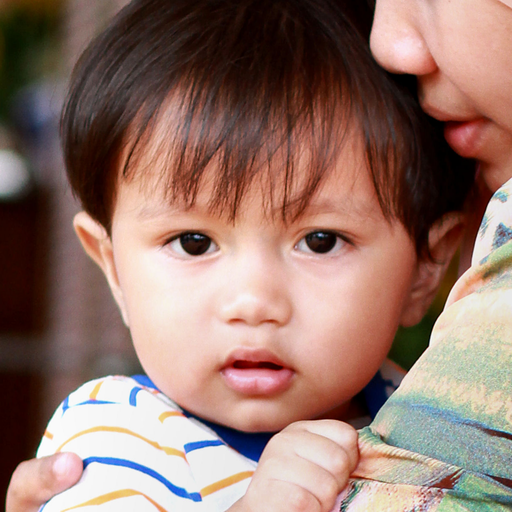}
    \end{subfigure}

    \begin{subfigure}[b]{0.19\textwidth}
        \includegraphics[width=\textwidth]{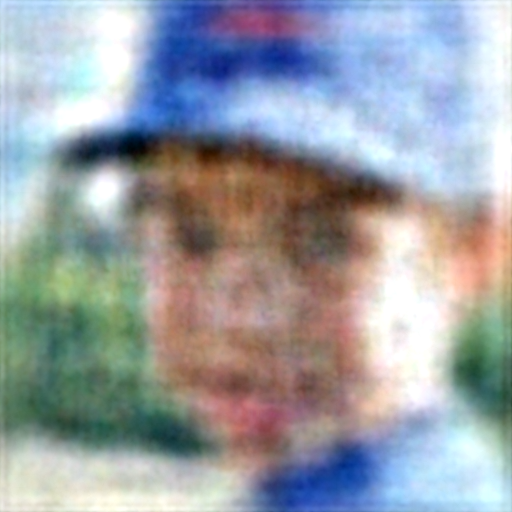}
    \end{subfigure}
    \begin{subfigure}[b]{0.19\textwidth}
        \includegraphics[width=\textwidth]{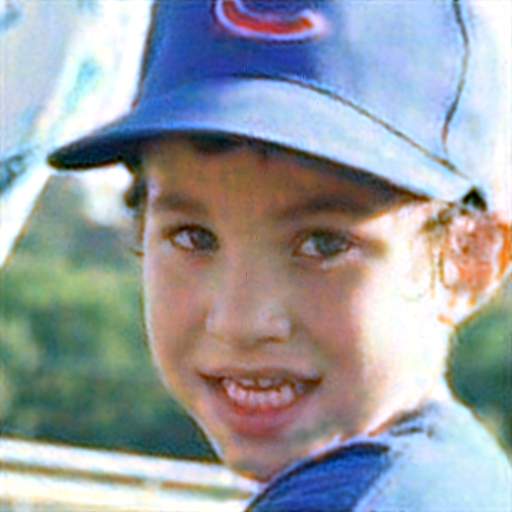}
    \end{subfigure}
    \begin{subfigure}[b]{0.19\textwidth}
        \includegraphics[width=\textwidth]{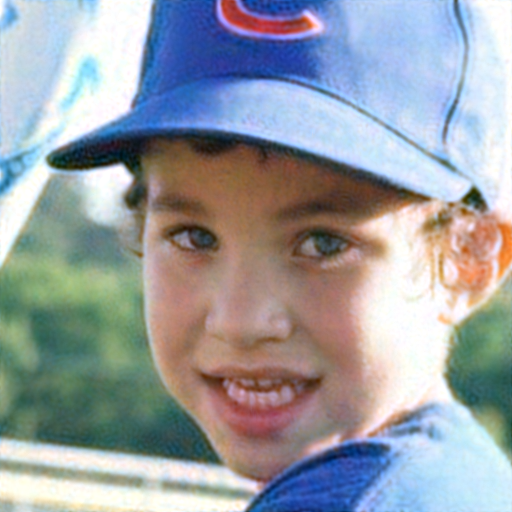}
    \end{subfigure}
    \begin{subfigure}[b]{0.19\textwidth}
        \includegraphics[width=\textwidth]{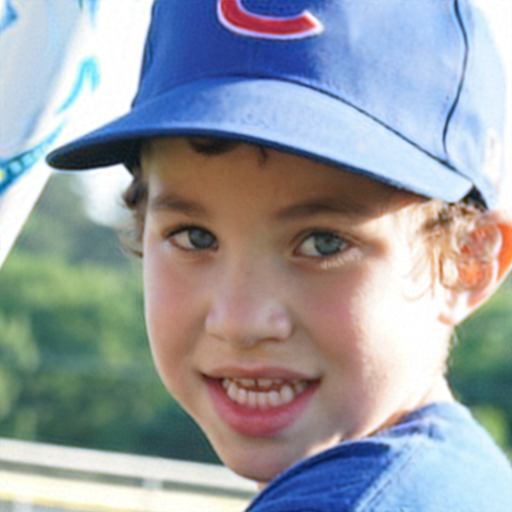}
    \end{subfigure}
    \begin{subfigure}[b]{0.19\textwidth}
        \includegraphics[width=\textwidth]{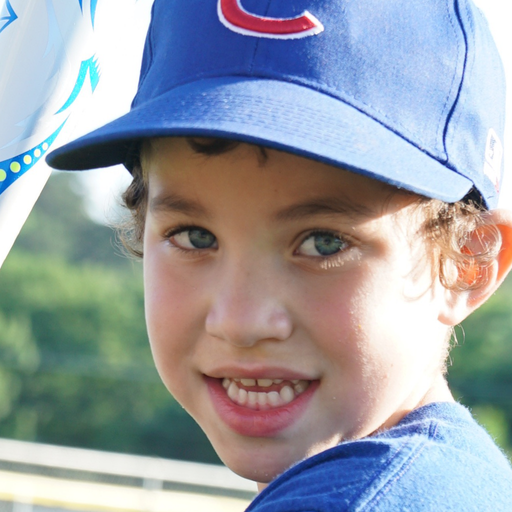}
    \end{subfigure}

    \begin{subfigure}[b]{0.19\textwidth}
        \includegraphics[width=\textwidth]{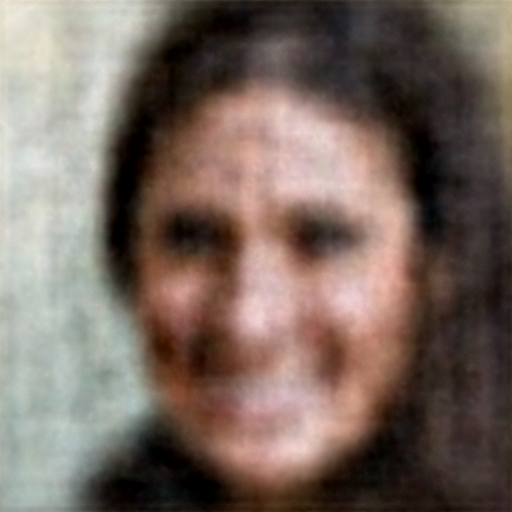}
    \end{subfigure}
    \begin{subfigure}[b]{0.19\textwidth}
        \includegraphics[width=\textwidth]{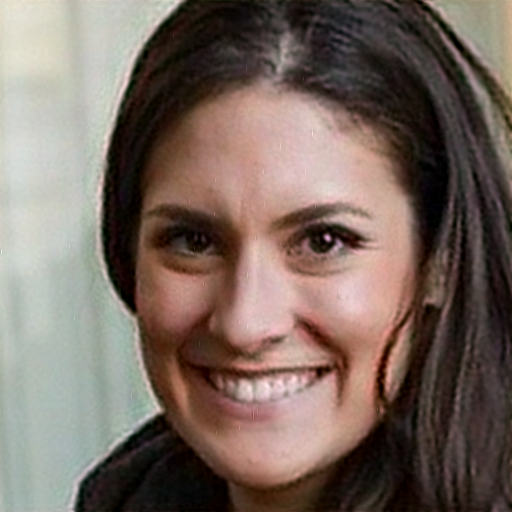}
    \end{subfigure}
    \begin{subfigure}[b]{0.19\textwidth}
        \includegraphics[width=\textwidth]{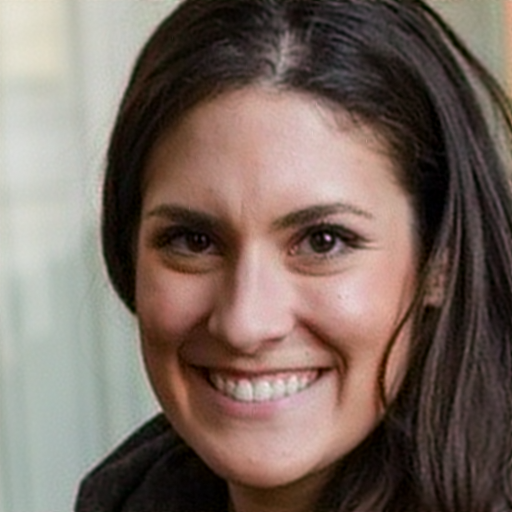}
    \end{subfigure}
    \begin{subfigure}[b]{0.19\textwidth}
        \includegraphics[width=\textwidth]{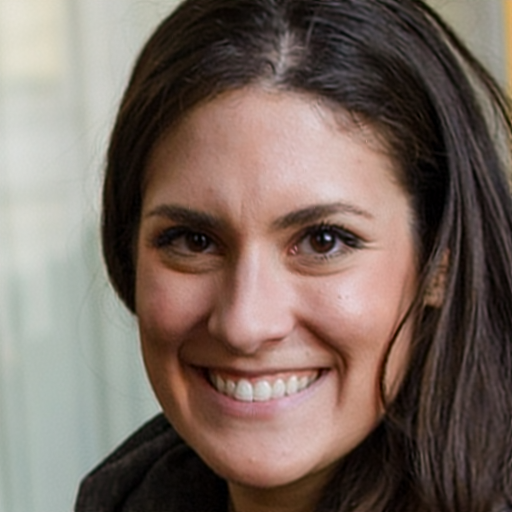}
    \end{subfigure}
    \begin{subfigure}[b]{0.19\textwidth}
        \includegraphics[width=\textwidth]{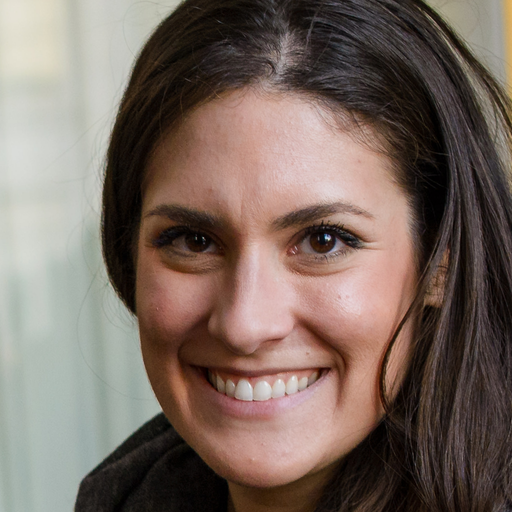}
    \end{subfigure}

    \begin{subfigure}[b]{0.19\textwidth}
        \includegraphics[width=\textwidth]{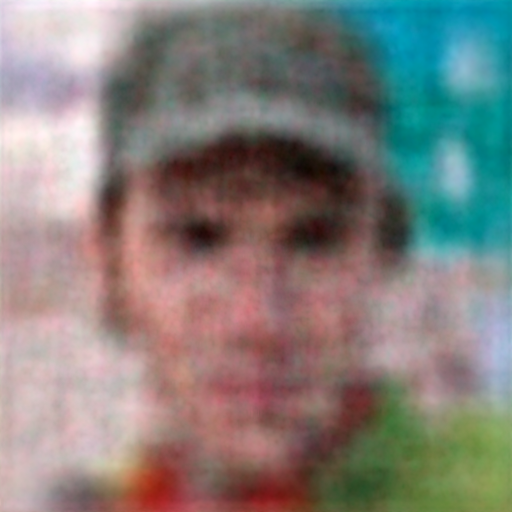}
    \end{subfigure}
    \begin{subfigure}[b]{0.19\textwidth}
        \includegraphics[width=\textwidth]{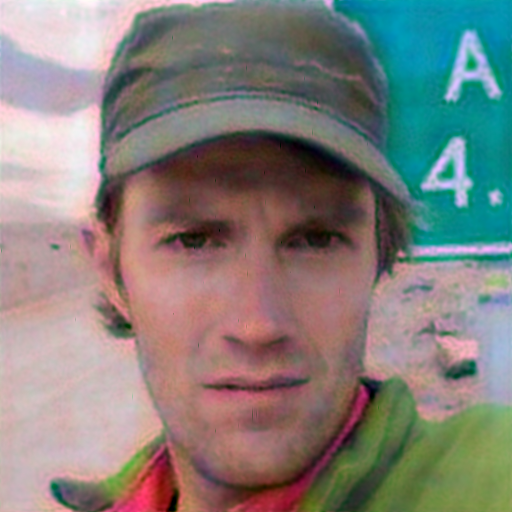}
    \end{subfigure}
    \begin{subfigure}[b]{0.19\textwidth}
        \includegraphics[width=\textwidth]{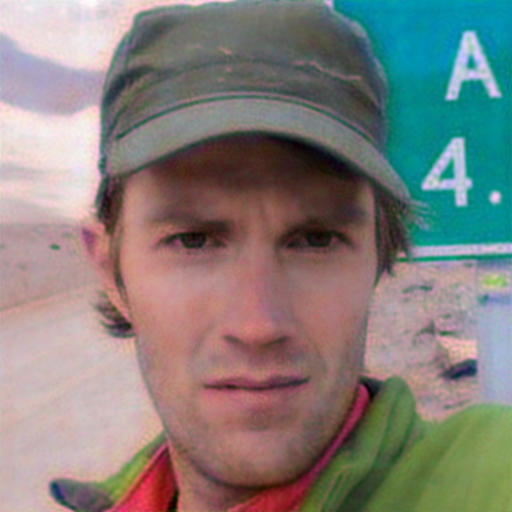}
    \end{subfigure}
    \begin{subfigure}[b]{0.19\textwidth}
        \includegraphics[width=\textwidth]{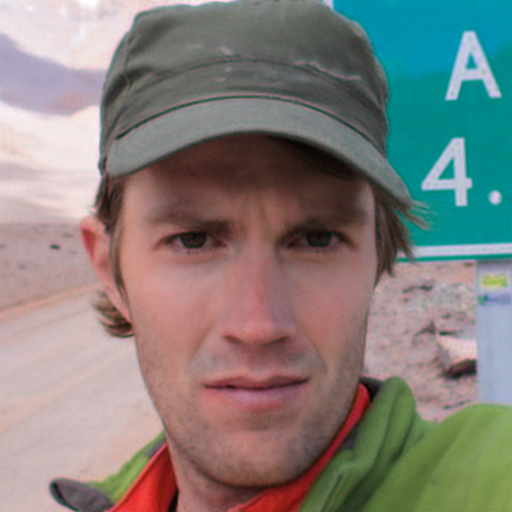}
    \end{subfigure}
    \begin{subfigure}[b]{0.19\textwidth}
        \includegraphics[width=\textwidth]{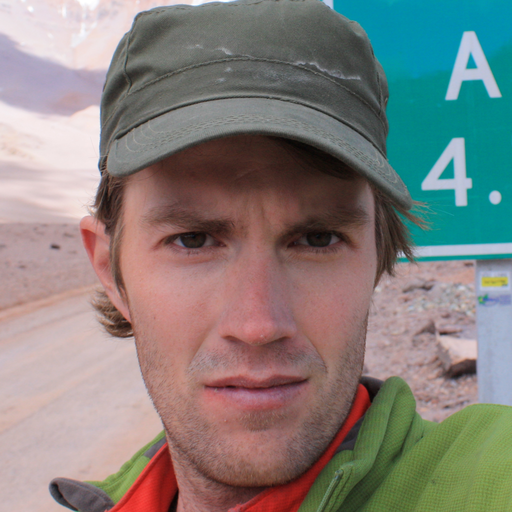}
    \end{subfigure}

    \begin{subfigure}[b]{0.19\textwidth}
        \includegraphics[width=\textwidth]{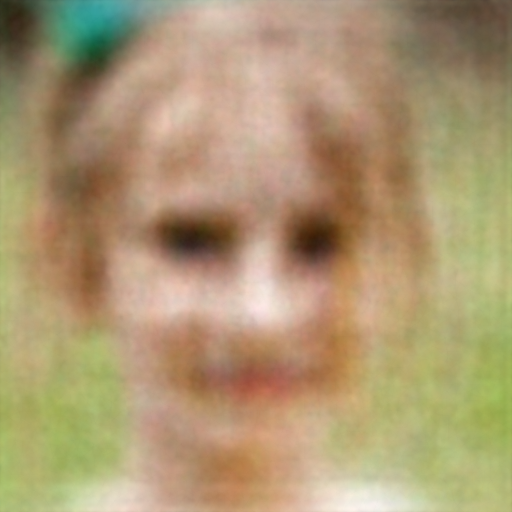}
    \end{subfigure}
    \begin{subfigure}[b]{0.19\textwidth}
        \includegraphics[width=\textwidth]{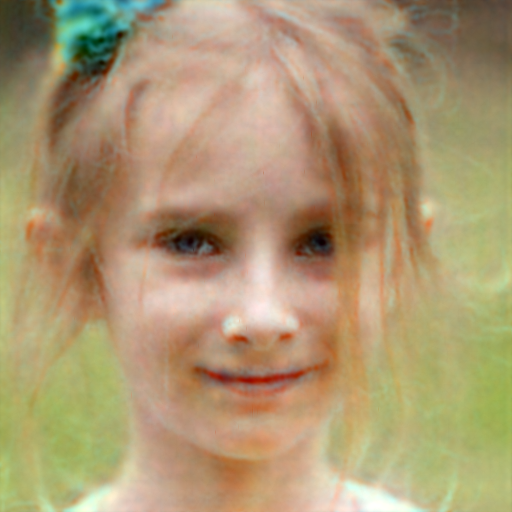}
    \end{subfigure}
    \begin{subfigure}[b]{0.19\textwidth}
        \includegraphics[width=\textwidth]{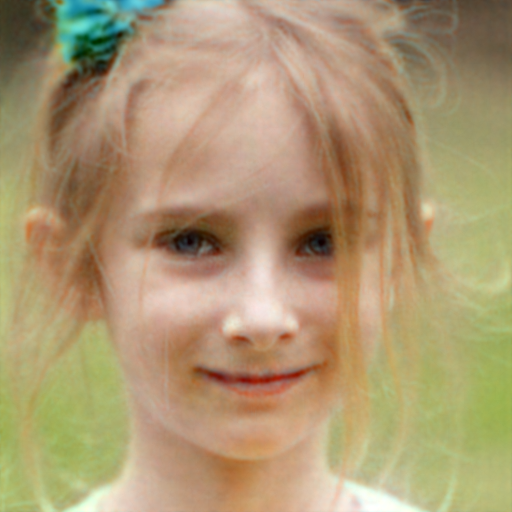}
    \end{subfigure}
    \begin{subfigure}[b]{0.19\textwidth}
        \includegraphics[width=\textwidth]{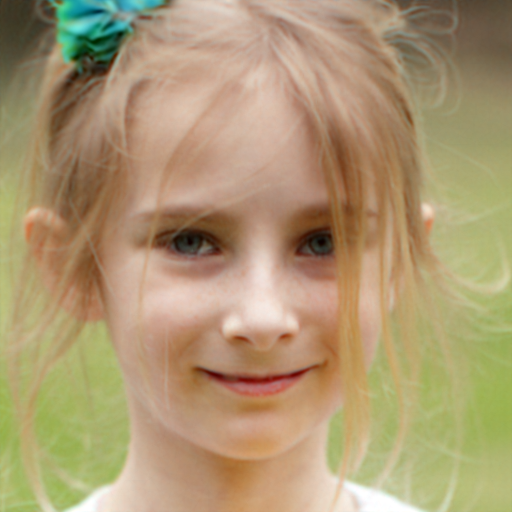}
    \end{subfigure}
    \begin{subfigure}[b]{0.19\textwidth}
        \includegraphics[width=\textwidth]{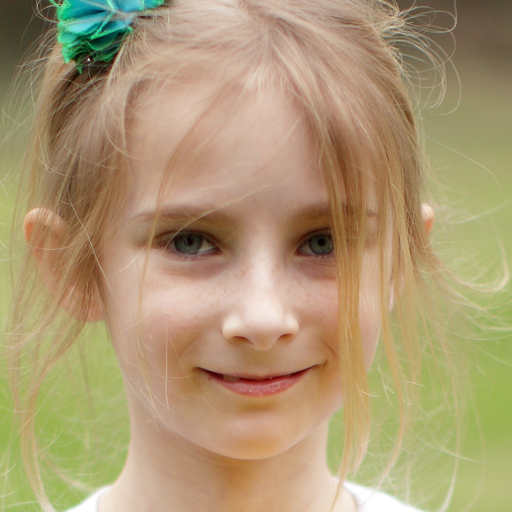}
    \end{subfigure}

    \caption{Qualitative results on the FFHQ dataset using ImageNet-trained EigenGS.}
    \label{fig:suppl_ffhq_vis}
\end{figure*}

\begin{figure*}
    \centering
    \begin{minipage}{0.19\textwidth}
        \centering
        ITER = 0
    \end{minipage}%
    \begin{minipage}{0.19\textwidth}
        \centering
        ITER = 10
    \end{minipage}%
    \begin{minipage}{0.19\textwidth}
        \centering
        ITER = 100
    \end{minipage}%
    \begin{minipage}{0.19\textwidth}
        \centering
        ITER = 1000
    \end{minipage}%
    \begin{minipage}{0.19\textwidth}
        \centering
        GT
    \end{minipage}
    \vspace{0.3em}

    \begin{subfigure}[b]{0.19\textwidth}
        \includegraphics[width=\textwidth]{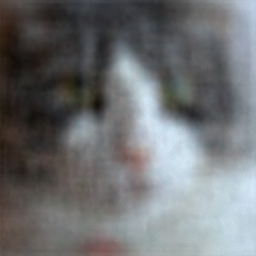}
    \end{subfigure}
    \begin{subfigure}[b]{0.19\textwidth}
        \includegraphics[width=\textwidth]{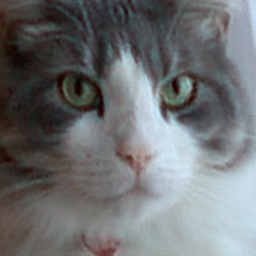}
    \end{subfigure}
    \begin{subfigure}[b]{0.19\textwidth}
        \includegraphics[width=\textwidth]{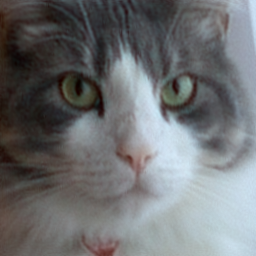}
    \end{subfigure}
    \begin{subfigure}[b]{0.19\textwidth}
        \includegraphics[width=\textwidth]{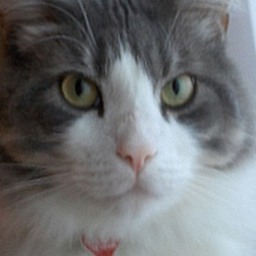}
    \end{subfigure}
    \begin{subfigure}[b]{0.19\textwidth}
        \includegraphics[width=\textwidth]{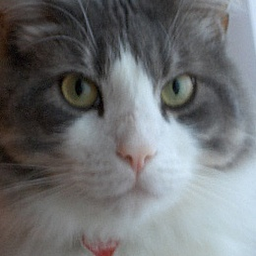}
    \end{subfigure}

    \begin{subfigure}[b]{0.19\textwidth}
        \includegraphics[width=\textwidth]{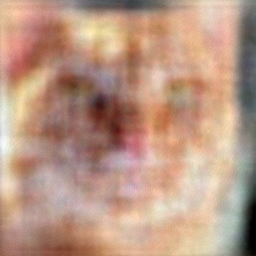}
    \end{subfigure}
    \begin{subfigure}[b]{0.19\textwidth}
        \includegraphics[width=\textwidth]{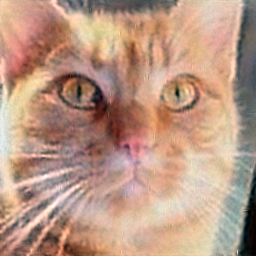}
    \end{subfigure}
    \begin{subfigure}[b]{0.19\textwidth}
        \includegraphics[width=\textwidth]{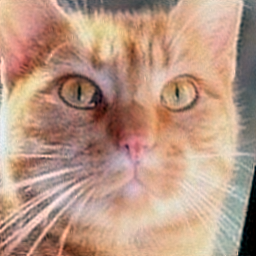}
    \end{subfigure}
    \begin{subfigure}[b]{0.19\textwidth}
        \includegraphics[width=\textwidth]{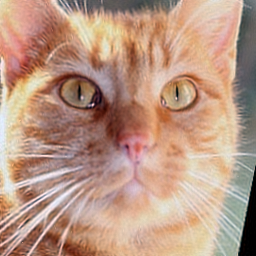}
    \end{subfigure}
    \begin{subfigure}[b]{0.19\textwidth}
        \includegraphics[width=\textwidth]{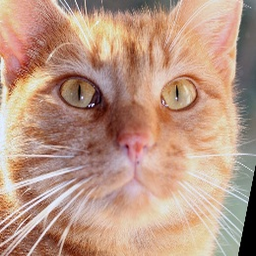}
    \end{subfigure}

    \begin{subfigure}[b]{0.19\textwidth}
        \includegraphics[width=\textwidth]{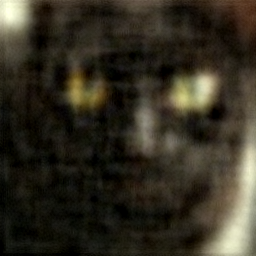}
    \end{subfigure}
    \begin{subfigure}[b]{0.19\textwidth}
        \includegraphics[width=\textwidth]{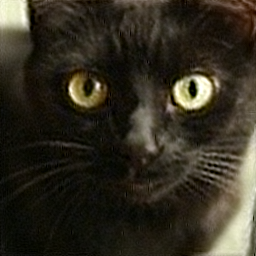}
    \end{subfigure}
    \begin{subfigure}[b]{0.19\textwidth}
        \includegraphics[width=\textwidth]{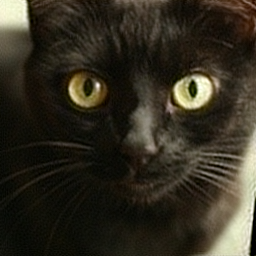}
    \end{subfigure}
    \begin{subfigure}[b]{0.19\textwidth}
        \includegraphics[width=\textwidth]{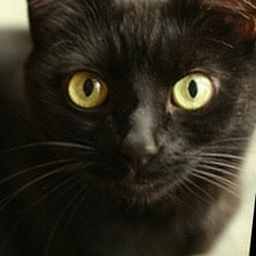}
    \end{subfigure}
    \begin{subfigure}[b]{0.19\textwidth}
        \includegraphics[width=\textwidth]{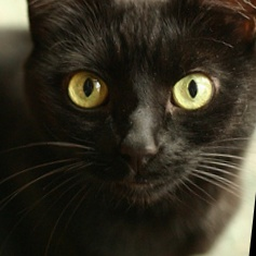}
    \end{subfigure}

    \begin{subfigure}[b]{0.19\textwidth}
        \includegraphics[width=\textwidth]{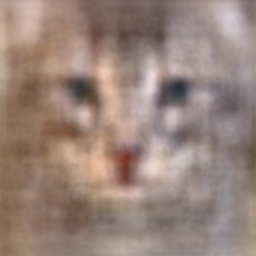}
    \end{subfigure}
    \begin{subfigure}[b]{0.19\textwidth}
        \includegraphics[width=\textwidth]{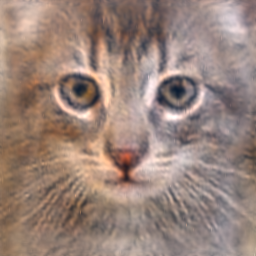}
    \end{subfigure}
    \begin{subfigure}[b]{0.19\textwidth}
        \includegraphics[width=\textwidth]{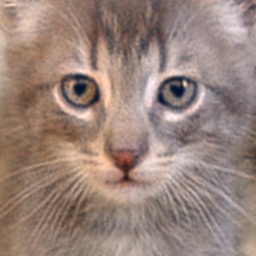}
    \end{subfigure}
    \begin{subfigure}[b]{0.19\textwidth}
        \includegraphics[width=\textwidth]{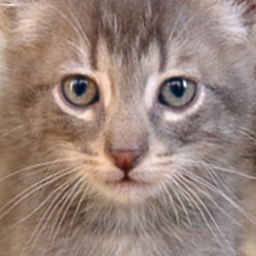}
    \end{subfigure}
    \begin{subfigure}[b]{0.19\textwidth}
        \includegraphics[width=\textwidth]{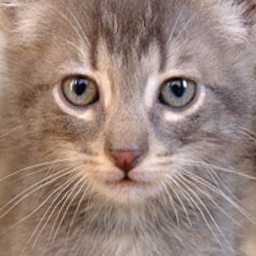}
    \end{subfigure}

    \begin{subfigure}[b]{0.19\textwidth}
        \includegraphics[width=\textwidth]{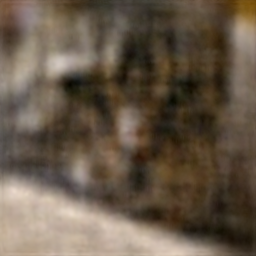}
    \end{subfigure}
    \begin{subfigure}[b]{0.19\textwidth}
        \includegraphics[width=\textwidth]{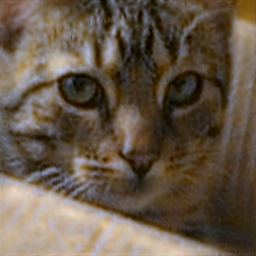}
    \end{subfigure}
    \begin{subfigure}[b]{0.19\textwidth}
        \includegraphics[width=\textwidth]{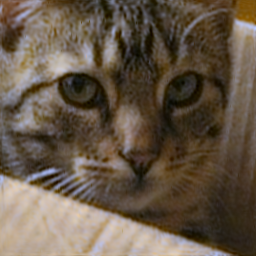}
    \end{subfigure}
    \begin{subfigure}[b]{0.19\textwidth}
        \includegraphics[width=\textwidth]{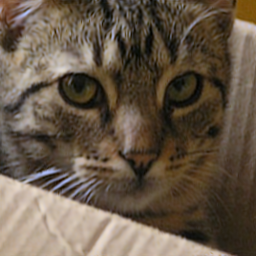}
    \end{subfigure}
    \begin{subfigure}[b]{0.19\textwidth}
        \includegraphics[width=\textwidth]{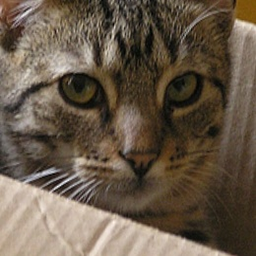}
    \end{subfigure}

    \caption{Qualitative results on the Cats dataset using ImageNet-trained EigenGS.}
    \label{fig:suppl_cats_vis}
\end{figure*}

\begin{figure*}
    \centering
    \begin{minipage}{0.19\textwidth}
        \centering
        ITER = 0
    \end{minipage}%
    \begin{minipage}{0.19\textwidth}
        \centering
        ITER = 10
    \end{minipage}%
    \begin{minipage}{0.19\textwidth}
        \centering
        ITER = 100
    \end{minipage}%
    \begin{minipage}{0.19\textwidth}
        \centering
        ITER = 1000
    \end{minipage}%
    \begin{minipage}{0.19\textwidth}
        \centering
        GT
    \end{minipage}
    \vspace{0.3em}

    \begin{subfigure}[b]{0.19\textwidth}
        \includegraphics[width=\textwidth]{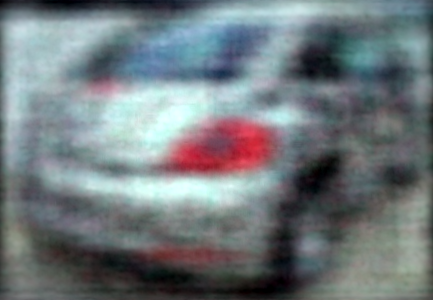}
    \end{subfigure}
    \begin{subfigure}[b]{0.19\textwidth}
        \includegraphics[width=\textwidth]{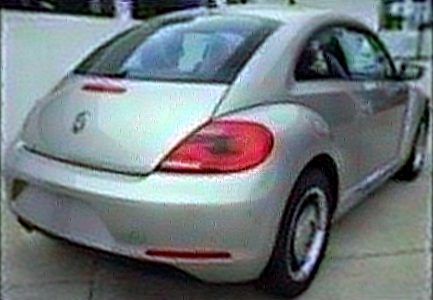}
    \end{subfigure}
    \begin{subfigure}[b]{0.19\textwidth}
        \includegraphics[width=\textwidth]{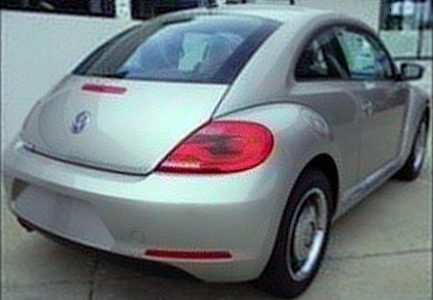}
    \end{subfigure}
    \begin{subfigure}[b]{0.19\textwidth}
        \includegraphics[width=\textwidth]{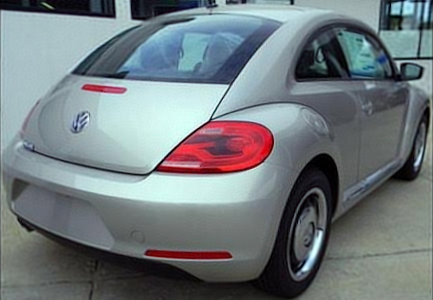}
    \end{subfigure}
    \begin{subfigure}[b]{0.19\textwidth}
        \includegraphics[width=\textwidth]{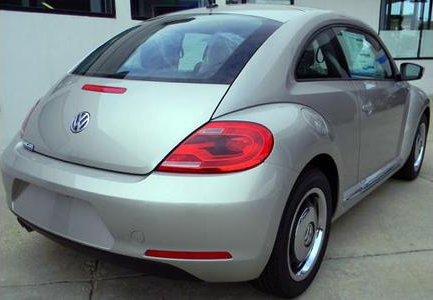}
    \end{subfigure}

    \begin{subfigure}[b]{0.19\textwidth}
        \includegraphics[width=\textwidth]{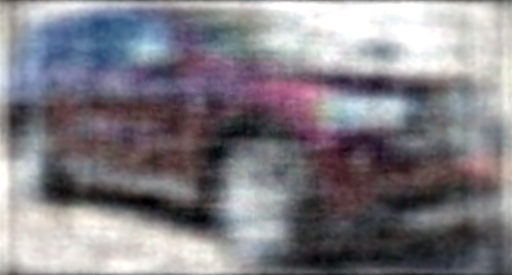}
    \end{subfigure}
    \begin{subfigure}[b]{0.19\textwidth}
        \includegraphics[width=\textwidth]{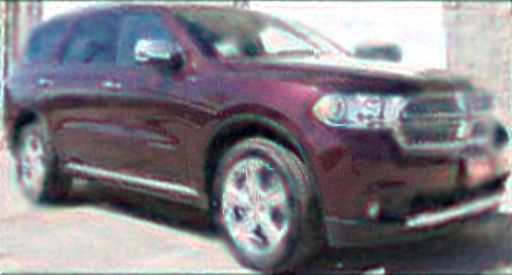}
    \end{subfigure}
    \begin{subfigure}[b]{0.19\textwidth}
        \includegraphics[width=\textwidth]{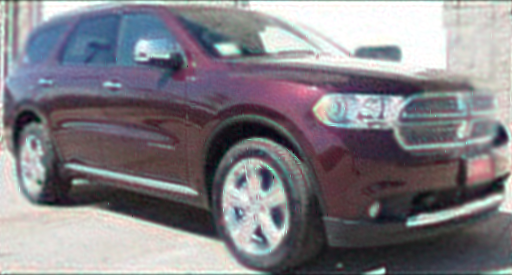}
    \end{subfigure}
    \begin{subfigure}[b]{0.19\textwidth}
        \includegraphics[width=\textwidth]{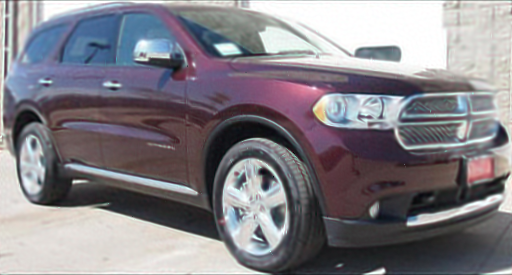}
    \end{subfigure}
    \begin{subfigure}[b]{0.19\textwidth}
        \includegraphics[width=\textwidth]{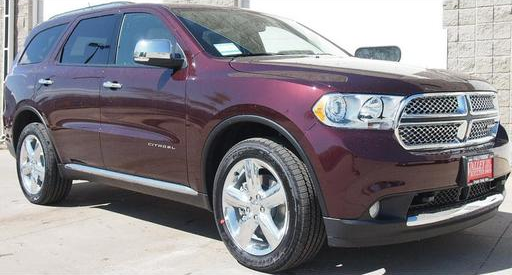}
    \end{subfigure}

    \begin{subfigure}[b]{0.19\textwidth}
        \includegraphics[width=\textwidth]{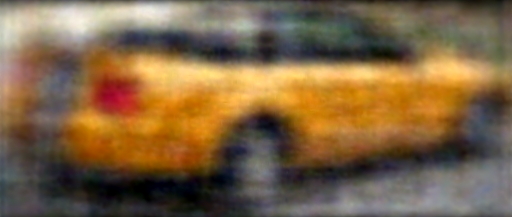}
    \end{subfigure}
    \begin{subfigure}[b]{0.19\textwidth}
        \includegraphics[width=\textwidth]{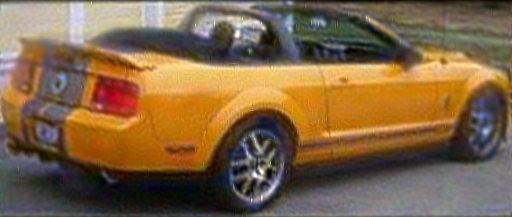}
    \end{subfigure}
    \begin{subfigure}[b]{0.19\textwidth}
        \includegraphics[width=\textwidth]{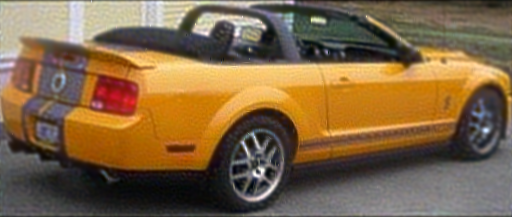}
    \end{subfigure}
    \begin{subfigure}[b]{0.19\textwidth}
        \includegraphics[width=\textwidth]{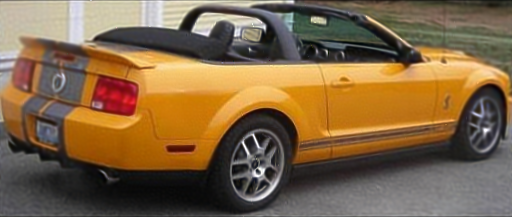}
    \end{subfigure}
    \begin{subfigure}[b]{0.19\textwidth}
        \includegraphics[width=\textwidth]{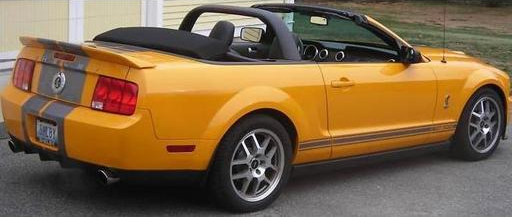}
    \end{subfigure}

    \begin{subfigure}[b]{0.19\textwidth}
        \includegraphics[width=\textwidth]{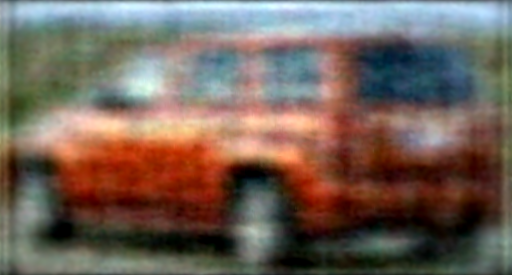}
    \end{subfigure}
    \begin{subfigure}[b]{0.19\textwidth}
        \includegraphics[width=\textwidth]{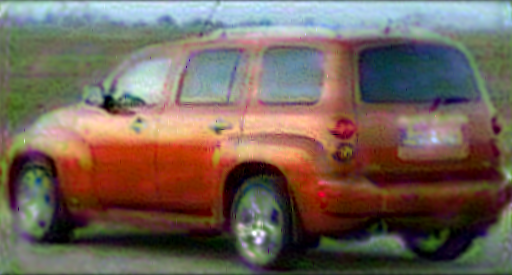}
    \end{subfigure}
    \begin{subfigure}[b]{0.19\textwidth}
        \includegraphics[width=\textwidth]{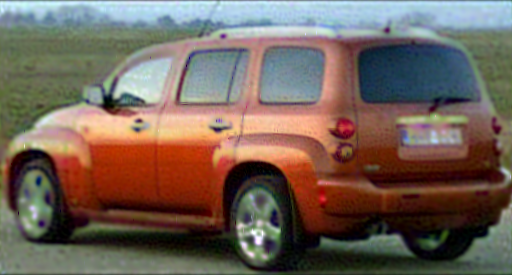}
    \end{subfigure}
    \begin{subfigure}[b]{0.19\textwidth}
        \includegraphics[width=\textwidth]{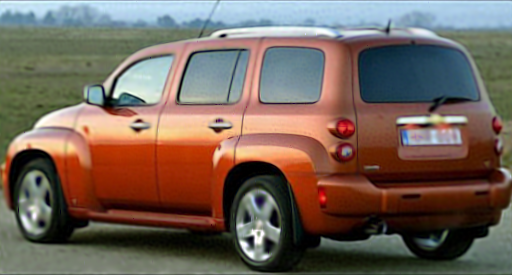}
    \end{subfigure}
    \begin{subfigure}[b]{0.19\textwidth}
        \includegraphics[width=\textwidth]{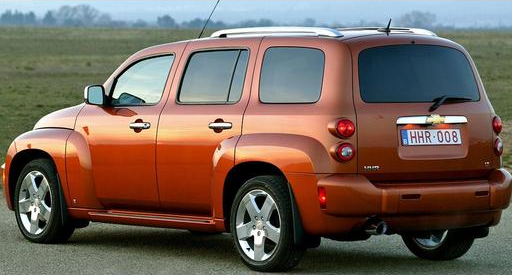}
    \end{subfigure}

    \begin{subfigure}[b]{0.19\textwidth}
        \includegraphics[width=\textwidth]{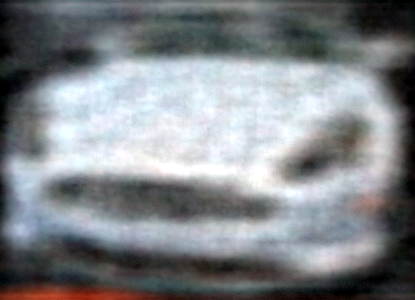}
    \end{subfigure}
    \begin{subfigure}[b]{0.19\textwidth}
        \includegraphics[width=\textwidth]{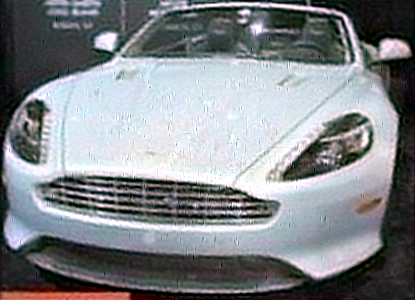}
    \end{subfigure}
    \begin{subfigure}[b]{0.19\textwidth}
        \includegraphics[width=\textwidth]{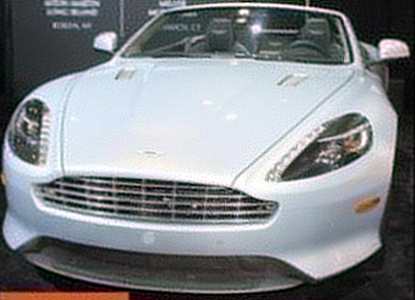}
    \end{subfigure}
    \begin{subfigure}[b]{0.19\textwidth}
        \includegraphics[width=\textwidth]{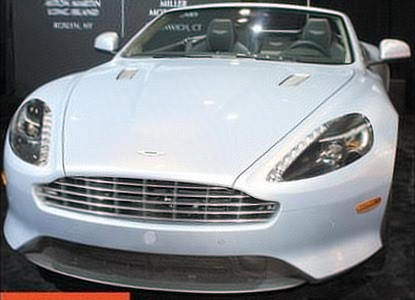}
    \end{subfigure}
    \begin{subfigure}[b]{0.19\textwidth}
        \includegraphics[width=\textwidth]{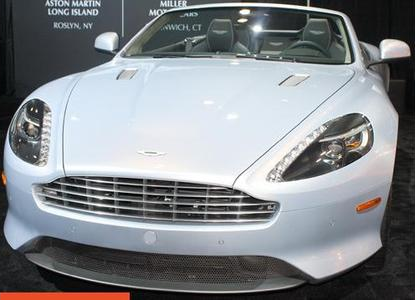}
    \end{subfigure}

    \caption{Qualitative results on the Cars dataset using ImageNet-trained EigenGS.}
    \label{fig:suppl_cars_vis}
\end{figure*}

We present comprehensive quantitative and qualitative results across our four benchmark datasets to show the effectiveness of the ImageNet-trained, universal EigenGS model. \Crefrange{fig:suppl_celeba_vis}{fig:suppl_cars_vis} show the visualization results using our ImageNet-trained EigenGS, which is noteworthy as it was trained on an unaligned collection of images with arbitrary sizes. The successful adaptation of this universal model to different datasets highlights the robustness of our approach.

\Crefrange{tab:suppl_celeba}{tab:suppl_cars} provide detailed performance metrics comparing three scenarios: the baseline GaussianImage method, dataset-specific trained EigenGS, and our ImageNet-trained universal EigenGS. In that, the dataset-specific trained EigenGS can represent the performance upper bound of our method, while the scores of ImageNet version corresponds to visualization. Overall, though the ImageNet-trained model shows a slight performance decrease compared to dataset-specific training, it consistently outperforms the baseline method across all datasets.

For CelebA, the reconstruction shows excellent preservation of facial features even at early iterations. Meanwhile, the FFHQ results demonstrate robust handling of high-resolution facial details. The Cats dataset shows our method's ability to capture broader structural elements. Lastly, despite the challenging variety of viewpoints in the Stanford Cars dataset, our method can still achieve high-quality reconstruction. These results particularly confirm that our EigenGS representation maintains consistent performance across different datasets even when using the ImageNet-trained model, which suggests that our approach successfully captures universal image statistics and generalizes well across domains.

\section{Spatial Frequency}
\label{sec:6.1}

\begin{figure*}
    \centering
    \begin{minipage}{0.19\textwidth}
        \centering
        $\mathcal{N}_h$
    \end{minipage}%
    \begin{minipage}{0.19\textwidth}
        \centering
        $\mathcal{N}_h$ (Ellipse) 
    \end{minipage}%
    \begin{minipage}{0.19\textwidth}
        \centering
        $\mathcal{N}_l$
    \end{minipage}%
    \begin{minipage}{0.19\textwidth}
        \centering
        $\mathcal{N}_l$ (Ellipse)
    \end{minipage}%
    \begin{minipage}{0.19\textwidth}
        \centering
        Ground Truth
    \end{minipage}
    \vspace{0.3em}

    \begin{subfigure}[b]{0.19\textwidth}
        \includegraphics[width=\textwidth]{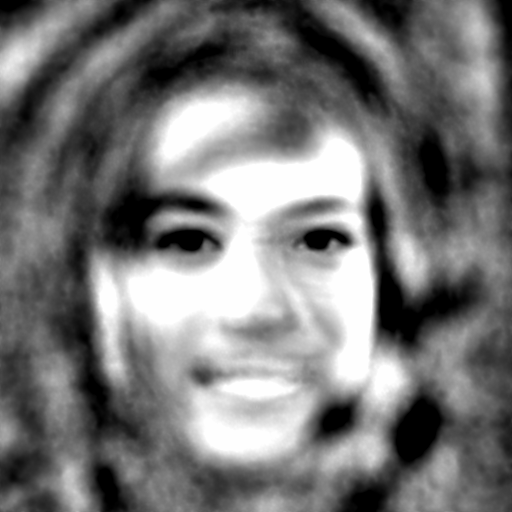}
    \end{subfigure}
    \begin{subfigure}[b]{0.19\textwidth}
        \includegraphics[width=\textwidth]{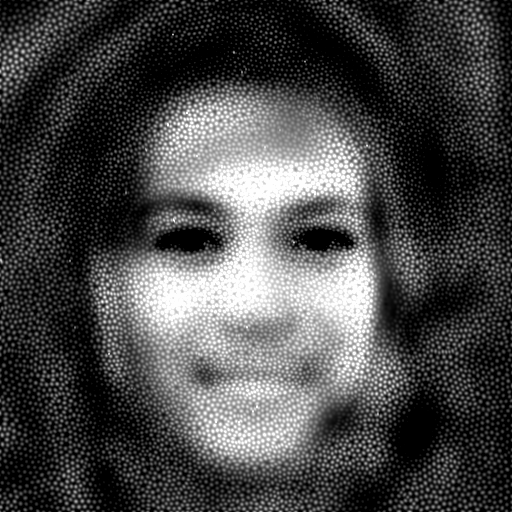}
    \end{subfigure}
    \begin{subfigure}[b]{0.19\textwidth}
        \includegraphics[width=\textwidth]{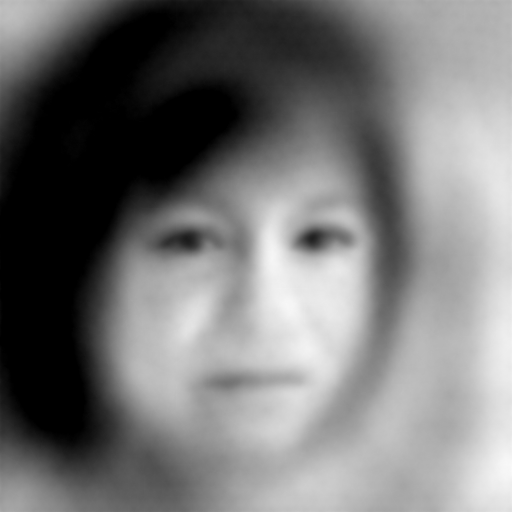}
    \end{subfigure}
    \begin{subfigure}[b]{0.19\textwidth}
        \includegraphics[width=\textwidth]{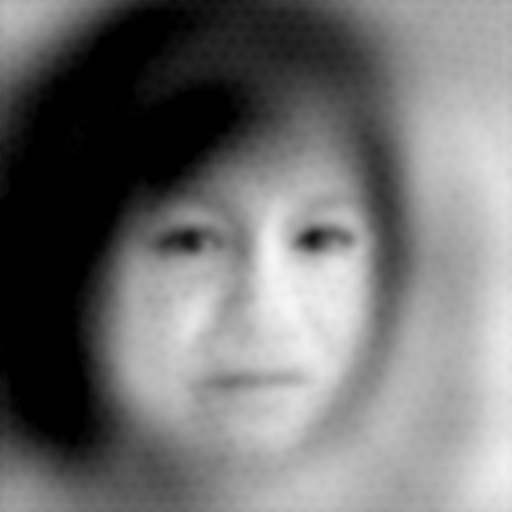}
    \end{subfigure}
    \begin{subfigure}[b]{0.19\textwidth}
        \includegraphics[width=\textwidth]{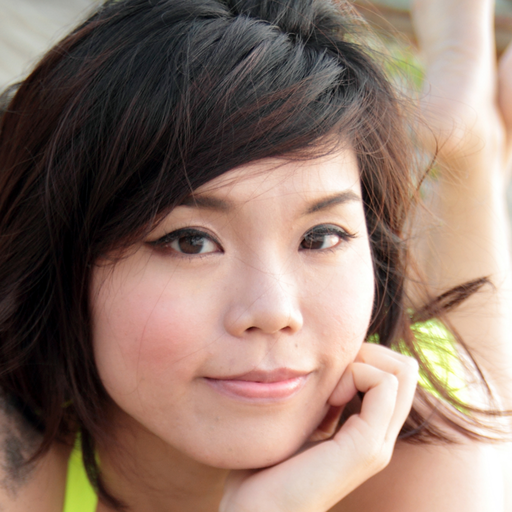}
    \end{subfigure}

    \begin{subfigure}[b]{0.19\textwidth}
        \includegraphics[width=\textwidth]{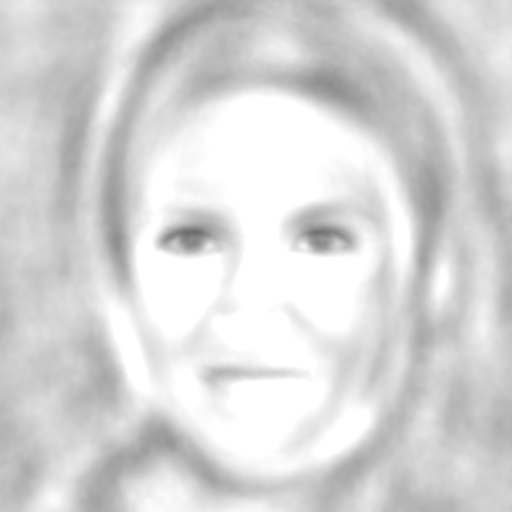}
    \end{subfigure}
    \begin{subfigure}[b]{0.19\textwidth}
        \includegraphics[width=\textwidth]{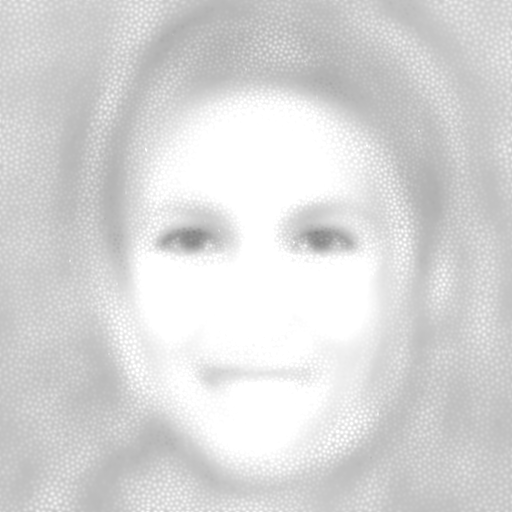}
    \end{subfigure}
    \begin{subfigure}[b]{0.19\textwidth}
        \includegraphics[width=\textwidth]{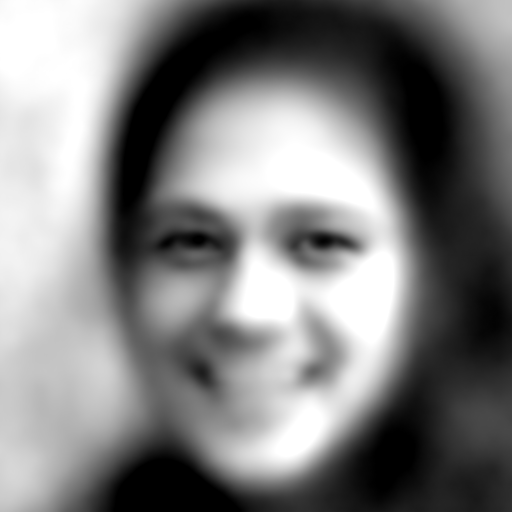}
    \end{subfigure}
    \begin{subfigure}[b]{0.19\textwidth}
        \includegraphics[width=\textwidth]{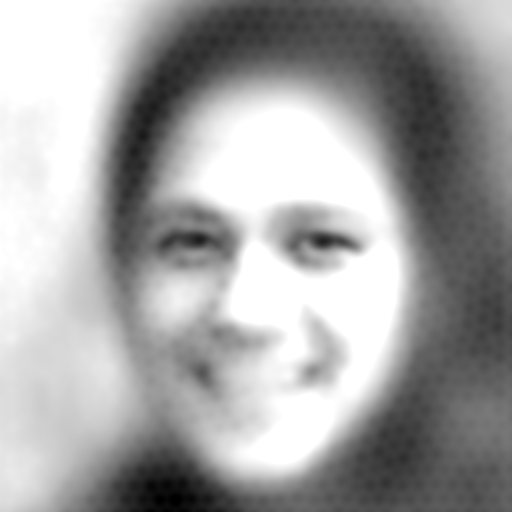}
    \end{subfigure}
    \begin{subfigure}[b]{0.19\textwidth}
        \includegraphics[width=\textwidth]{suppl/ffhq_vis/074.png}
    \end{subfigure}

    \begin{subfigure}[b]{0.19\textwidth}
        \includegraphics[width=\textwidth]{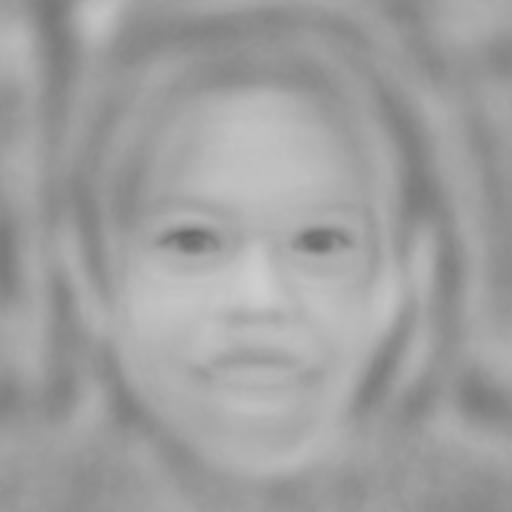}
    \end{subfigure}
    \begin{subfigure}[b]{0.19\textwidth}
        \includegraphics[width=\textwidth]{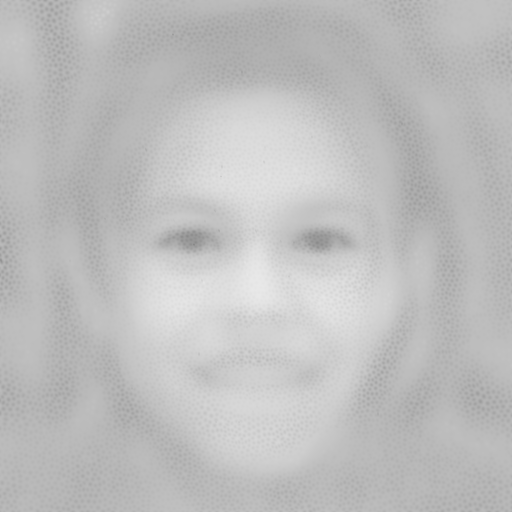}
    \end{subfigure}
    \begin{subfigure}[b]{0.19\textwidth}
        \includegraphics[width=\textwidth]{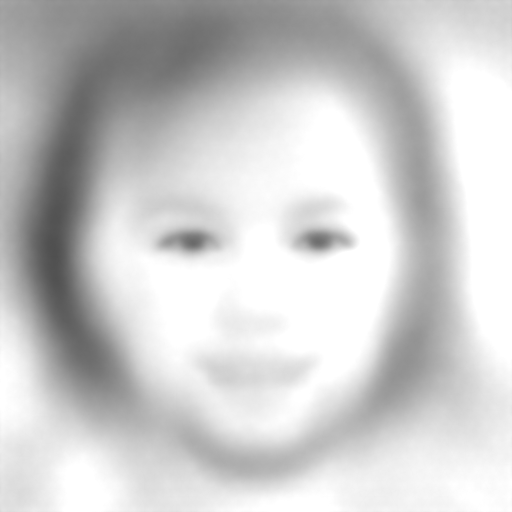}
    \end{subfigure}
    \begin{subfigure}[b]{0.19\textwidth}
        \includegraphics[width=\textwidth]{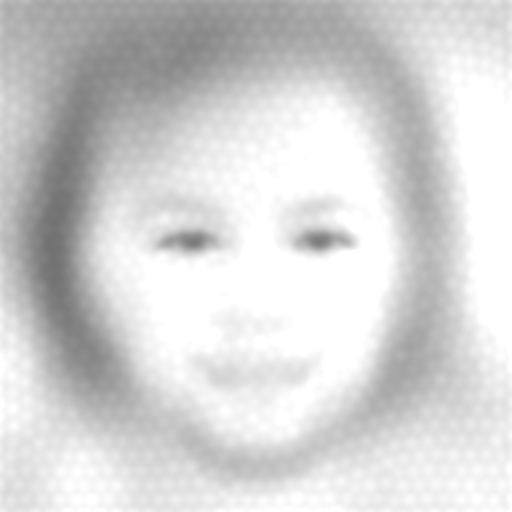}
    \end{subfigure}
    \begin{subfigure}[b]{0.19\textwidth}
        \includegraphics[width=\textwidth]{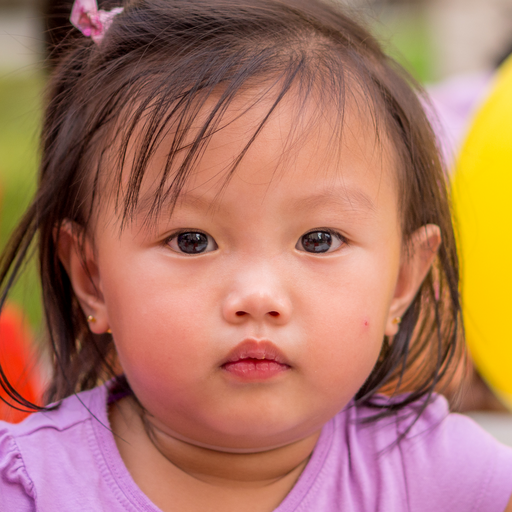}
    \end{subfigure}

    \begin{subfigure}[b]{0.19\textwidth}
        \includegraphics[width=\textwidth]{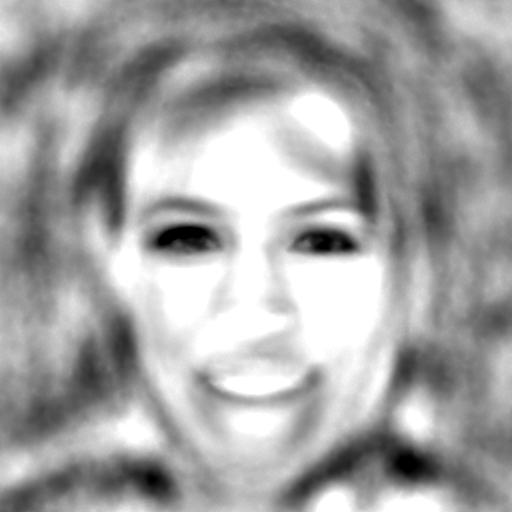}
    \end{subfigure}
    \begin{subfigure}[b]{0.19\textwidth}
        \includegraphics[width=\textwidth]{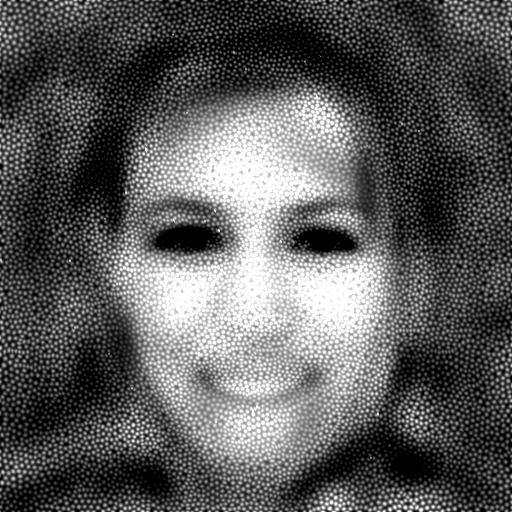}
    \end{subfigure}
    \begin{subfigure}[b]{0.19\textwidth}
        \includegraphics[width=\textwidth]{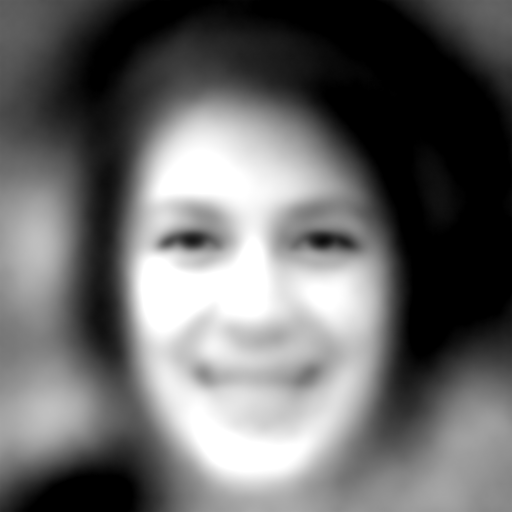}
    \end{subfigure}
    \begin{subfigure}[b]{0.19\textwidth}
        \includegraphics[width=\textwidth]{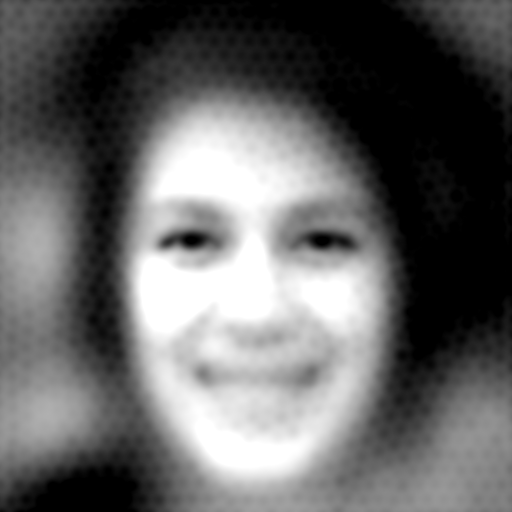}
    \end{subfigure}
    \begin{subfigure}[b]{0.19\textwidth}
        \includegraphics[width=\textwidth]{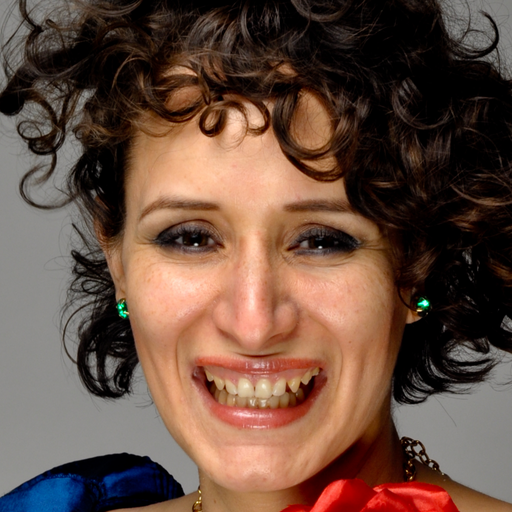}
    \end{subfigure}

    \begin{subfigure}[b]{0.19\textwidth}
        \includegraphics[width=\textwidth]{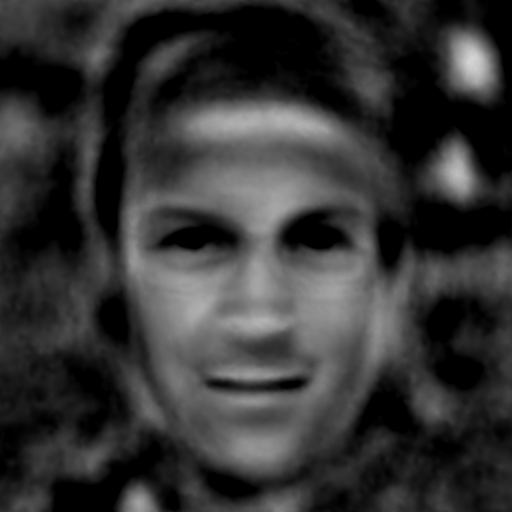}
    \end{subfigure}
    \begin{subfigure}[b]{0.19\textwidth}
        \includegraphics[width=\textwidth]{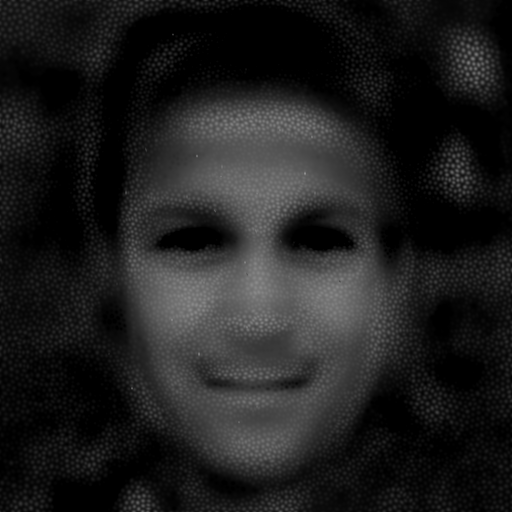}
    \end{subfigure}
    \begin{subfigure}[b]{0.19\textwidth}
        \includegraphics[width=\textwidth]{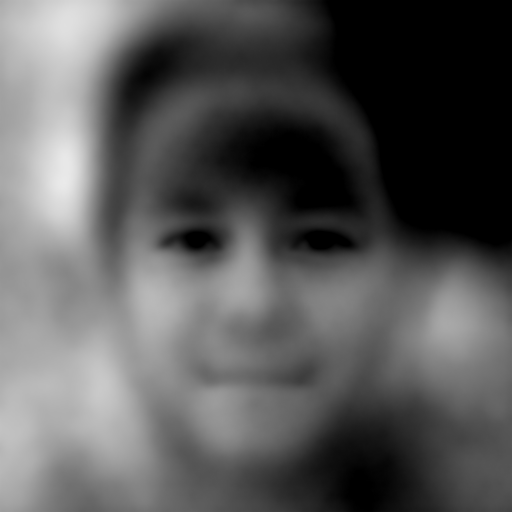}
    \end{subfigure}
    \begin{subfigure}[b]{0.19\textwidth}
        \includegraphics[width=\textwidth]{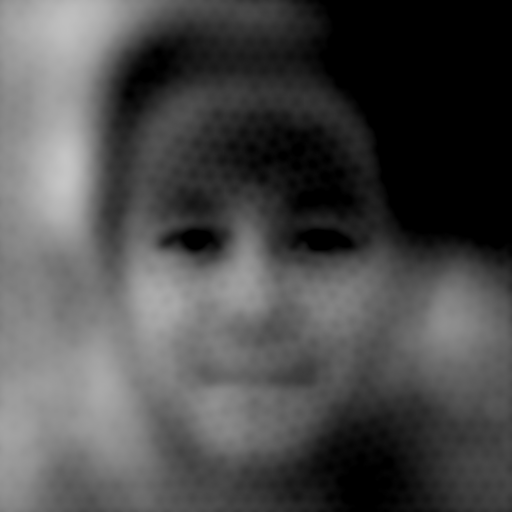}
    \end{subfigure}
    \begin{subfigure}[b]{0.19\textwidth}
        \includegraphics[width=\textwidth]{suppl/ffhq_vis/078.png}
    \end{subfigure}

    \caption{Spatial frequency separation using normalized Y-channel rendering, where darker and brighter regions represent negative and positive Gaussian clusters respectively. From left to right: $\mathcal{N}_h$ shows reconstruction from only high-frequency Gaussians; $\mathcal{N}_h$ (Ellipse) visualizes the smaller spatial coverage of detail-oriented Gaussians; $\mathcal{N}_l$ shows reconstruction using only low-frequency Gaussians, $\mathcal{N}_l$ (Ellipse) displays the same Gaussians with reduced scale to highlight their spatial extent, and GT shows the ground truth image.}
    \label{fig:freq_vis}
\end{figure*}

Our frequency-aware learning mechanism reveals interesting characteristics in how different spatial frequencies are handled in image reconstruction. Through visualization of the Gaussian components shown in \Cref{fig:freq_vis}, we can observe the larger spatial extent of the low-frequency Gaussians, $\mathcal{N}_l$, indicating their role in capturing larger structural elements. In contrast, the high-frequency set $\mathcal{N}_h$ consists of smaller Gaussians that provide fine detail refinement, as evidenced by their more compact elliptical boundaries.

The low-frequency set $\mathcal{N}_l$ comprises approximately 10\% of the overall number of Gaussians, which establishes a foundational structure that remains largely stable during optimization. This foundation serves as a ground layer upon which high-frequency details are subsequently refined. This hierarchical approach proves particularly effective in preventing ``penny-round-tile" artifacts, which typically emerge when all Gaussians converge to uniformly small sizes and locate side-by-side on the image to create a visually distracting pattern, as shown in the main paper.

\section{Color Space}
\label{sec:6.2}

\begin{figure*}
    \centering
    
    \begin{minipage}{0.19\textwidth}
        \centering
        ITER = 0
    \end{minipage}%
    \begin{minipage}{0.19\textwidth}
        \centering
        ITER = 10
    \end{minipage}%
    \begin{minipage}{0.19\textwidth}
        \centering
        ITER = 100
    \end{minipage}
    \begin{minipage}{0.19\textwidth}
        \centering
        ITER = 1000
    \end{minipage}
    \begin{minipage}{0.19\textwidth}
        \centering
        GT
    \end{minipage}
    \vspace{0.3em}

    \begin{subfigure}[b]{0.19\textwidth}
        \includegraphics[width=\textwidth]{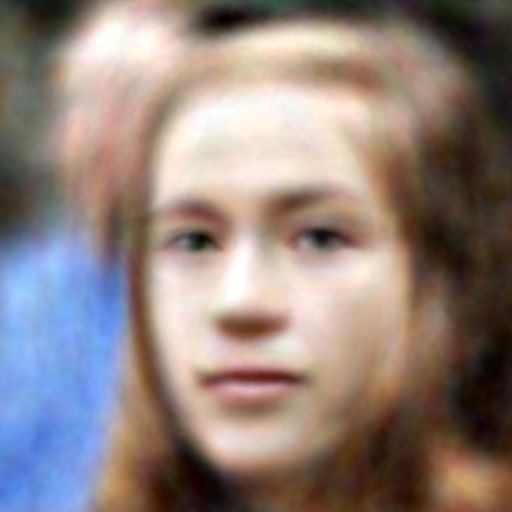}
    \end{subfigure}
    \begin{subfigure}[b]{0.19\textwidth}
        \includegraphics[width=\textwidth]{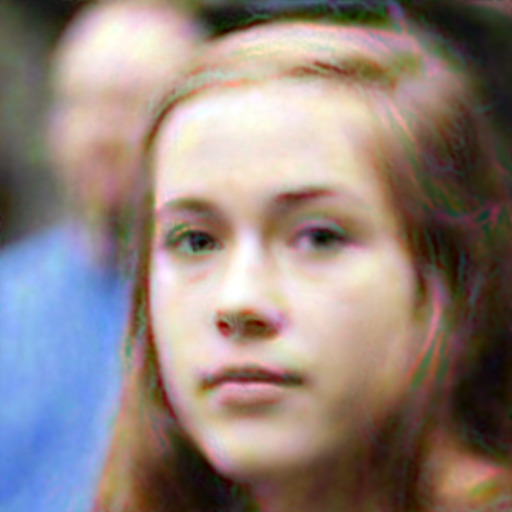}
    \end{subfigure}
    \begin{subfigure}[b]{0.19\textwidth}
        \includegraphics[width=\textwidth]{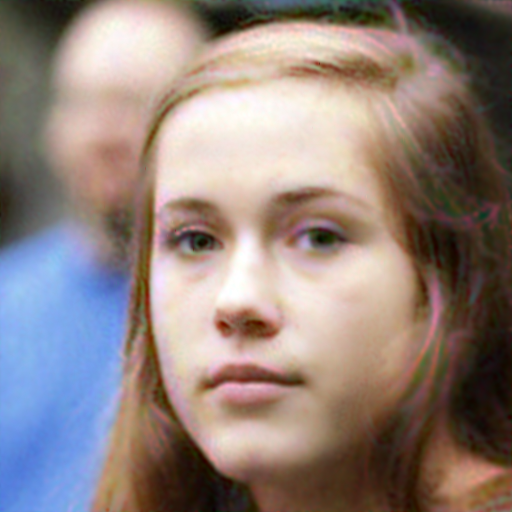}
    \end{subfigure}
    \begin{subfigure}[b]{0.19\textwidth}
        \includegraphics[width=\textwidth]{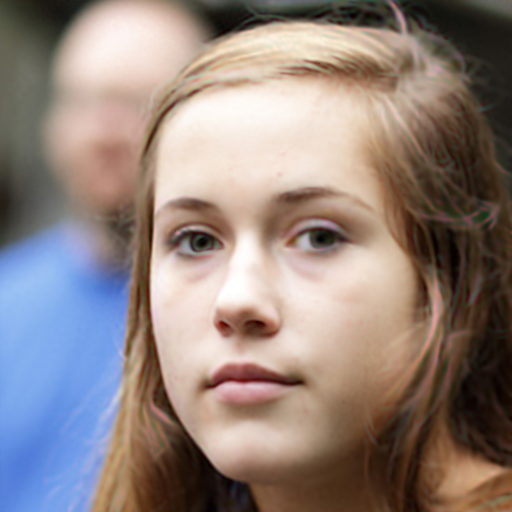}
    \end{subfigure}
    \begin{subfigure}[b]{0.19\textwidth}
        \includegraphics[width=\textwidth]{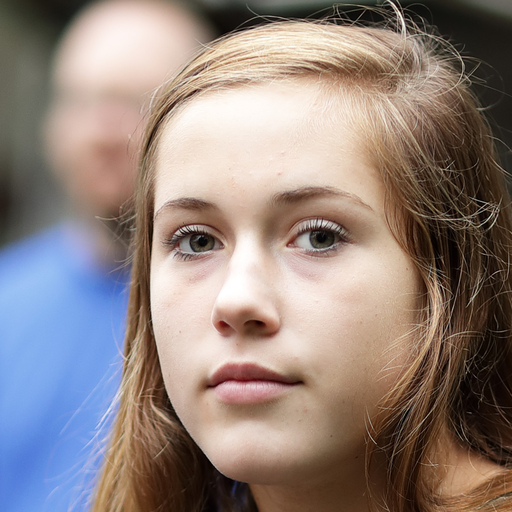}
    \end{subfigure}

    \begin{subfigure}[b]{0.19\textwidth}
        \includegraphics[width=\textwidth]{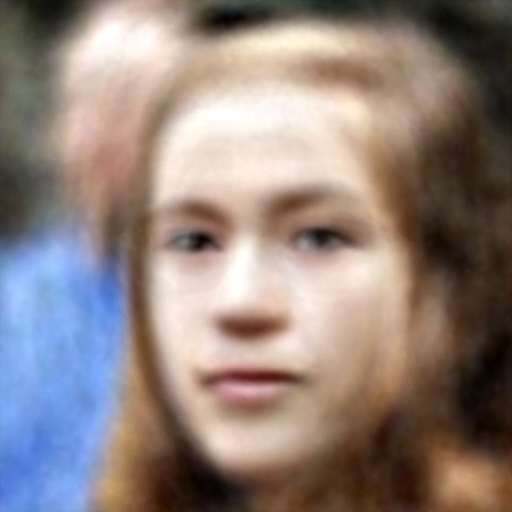}
    \end{subfigure}
    \begin{subfigure}[b]{0.19\textwidth}
        \includegraphics[width=\textwidth]{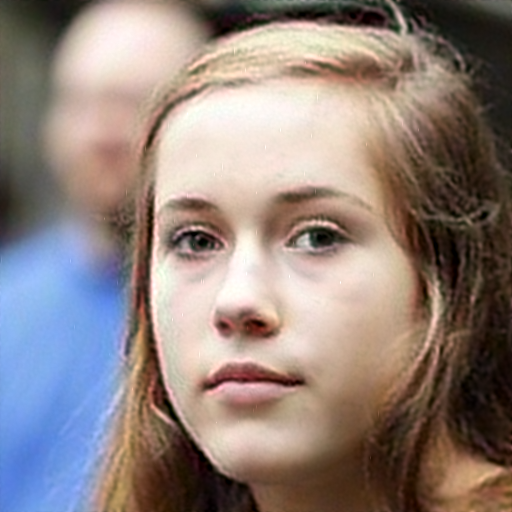}
    \end{subfigure}
    \begin{subfigure}[b]{0.19\textwidth}
        \includegraphics[width=\textwidth]{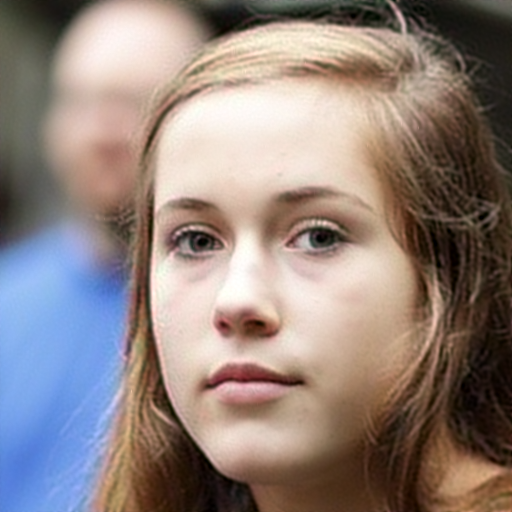}
    \end{subfigure}
    \begin{subfigure}[b]{0.19\textwidth}
        \includegraphics[width=\textwidth]{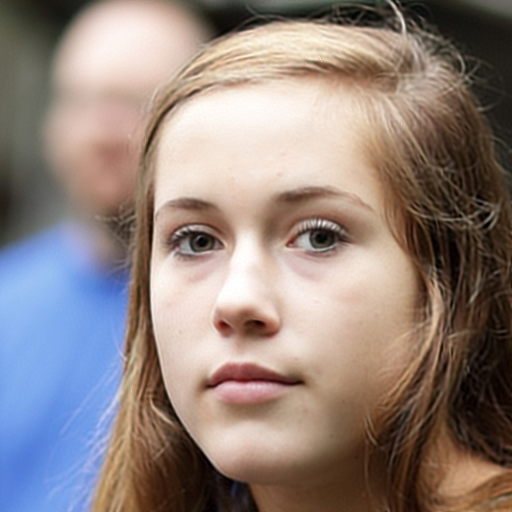}
    \end{subfigure}
    \begin{subfigure}[b]{0.19\textwidth}
        \includegraphics[width=\textwidth]{suppl/ffhq_vis/072.png}
    \end{subfigure}

    \caption{Color space comparison on FFHQ dataset. Top row: RGB color space reconstruction shows noticeable color shifts particularly at early iteration, \eg, the red and blue fringe-like artifacts around the hair and eyes. Bottom row: YCbCr color space reconstruction shows superior early convergence with sharp detail preservation, even for the hair regions, exhibiting minimal color artifacts throughout the optimization process.}
    \label{fig:colorspace_test72}
\end{figure*}

\begin{figure*}
    \centering
    
    \begin{minipage}{0.19\textwidth}
        \centering
        ITER = 0
    \end{minipage}%
    \begin{minipage}{0.19\textwidth}
        \centering
        ITER = 10
    \end{minipage}%
    \begin{minipage}{0.19\textwidth}
        \centering
        ITER = 100
    \end{minipage}
    \begin{minipage}{0.19\textwidth}
        \centering
        ITER = 1000
    \end{minipage}
    \begin{minipage}{0.19\textwidth}
        \centering
        GT
    \end{minipage}
    \vspace{0.3em}

    \begin{subfigure}[b]{0.19\textwidth}
        \includegraphics[width=\textwidth]{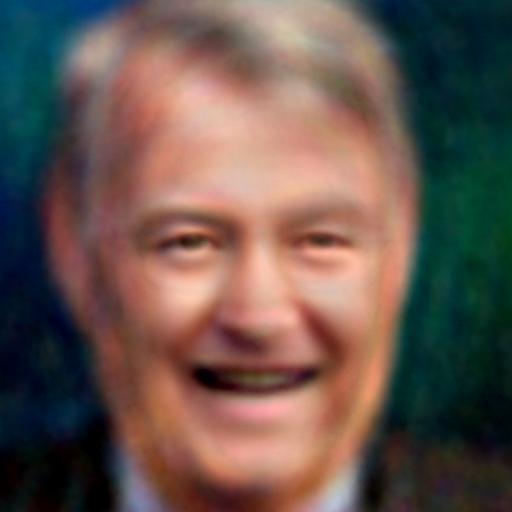}
    \end{subfigure}
    \begin{subfigure}[b]{0.19\textwidth}
        \includegraphics[width=\textwidth]{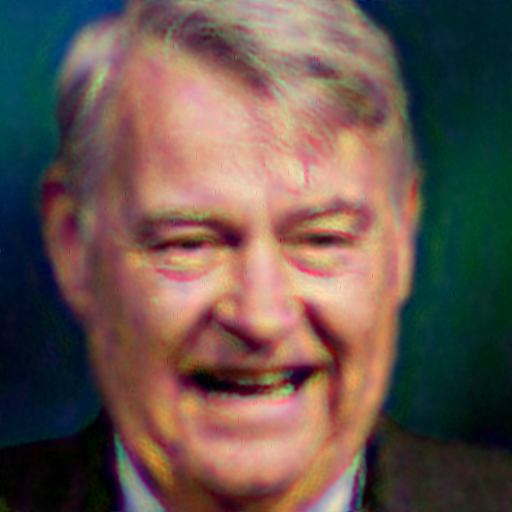}
    \end{subfigure}
    \begin{subfigure}[b]{0.19\textwidth}
        \includegraphics[width=\textwidth]{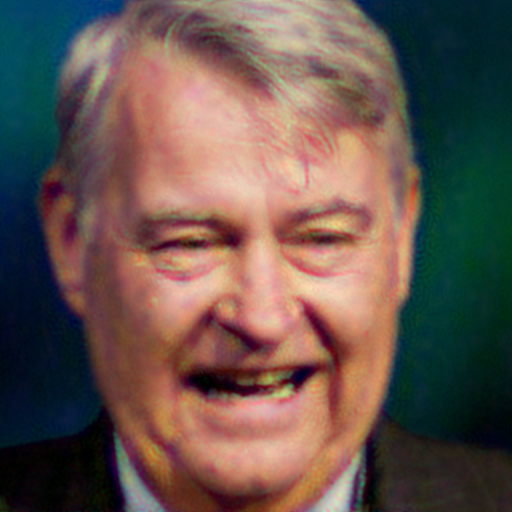}
    \end{subfigure}
    \begin{subfigure}[b]{0.19\textwidth}
        \includegraphics[width=\textwidth]{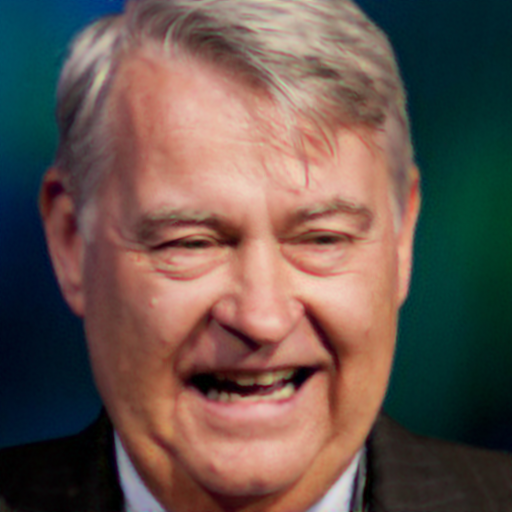}
    \end{subfigure}
    \begin{subfigure}[b]{0.19\textwidth}
        \includegraphics[width=\textwidth]{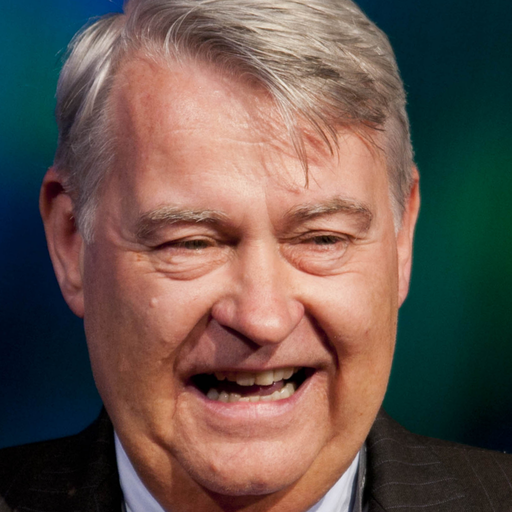}
    \end{subfigure}

    \begin{subfigure}[b]{0.19\textwidth}
        \includegraphics[width=\textwidth]{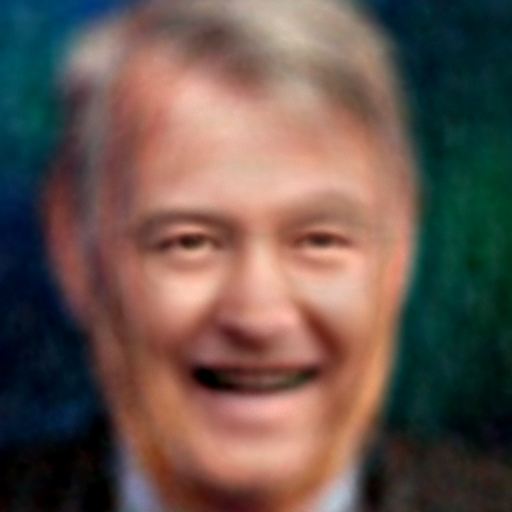}
    \end{subfigure}
    \begin{subfigure}[b]{0.19\textwidth}
        \includegraphics[width=\textwidth]{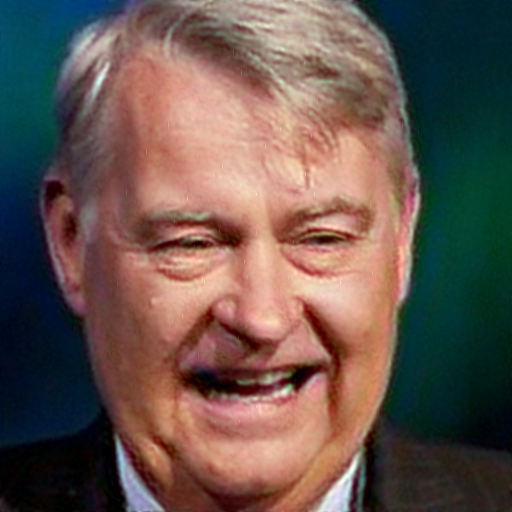}
    \end{subfigure}
    \begin{subfigure}[b]{0.19\textwidth}
        \includegraphics[width=\textwidth]{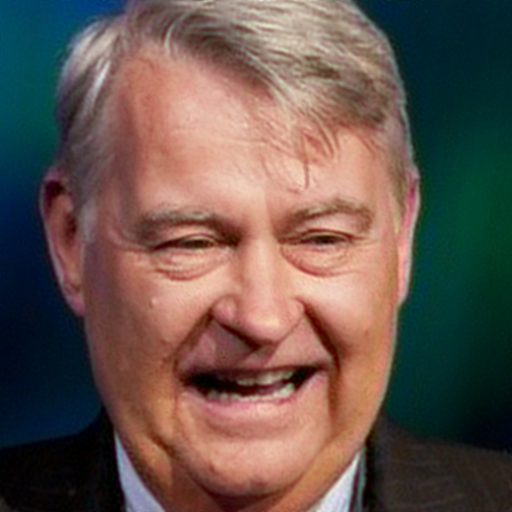}
    \end{subfigure}
    \begin{subfigure}[b]{0.19\textwidth}
        \includegraphics[width=\textwidth]{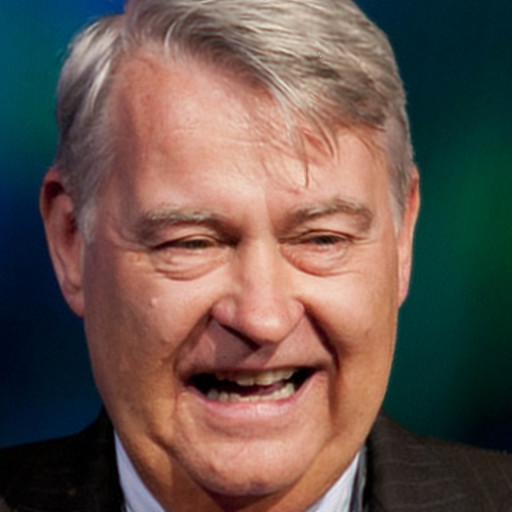}
    \end{subfigure}
    \begin{subfigure}[b]{0.19\textwidth}
        \includegraphics[width=\textwidth]{suppl/ffhq_vis/047.png}
    \end{subfigure}

    \caption{Another example of color space comparison on FFHQ dataset. Top: RGB color space reconstruction displays visible color bleeding during intermediate iterations. Bottom: YCbCr color space reconstruction maintains better color fidelity and stability, further supporting the advantages of separated luminance-chrominance processing in our method.}
    \label{fig:colorspace_test47}
\end{figure*}

Our experimental results show that the choice of color space may impact both the visual quality and convergence characteristics of EigenGS. By visualizing the reconstruction process
\Crefrange{fig:colorspace_test72}{fig:colorspace_test47}, we observe distinct differences between RGB and YCbCr color space implementations, particularly in handling PCA-based reconstruction artifacts.

The YCbCr color space offers natural advantages through its separation of luminance (Y: 16-235) and chrominance (Cb/Cr: 16-240) components. This design provides inherent margins for handling reconstruction values that may fall outside the typical 0-255 range during PCA-based processing. In contrast, when using RGB color space, all three channels are processed with equal weighting, and thus, the reconstruction is more susceptible to outliers, particularly in regions with extreme values.

A particularly notable advantage of YCbCr implementation is its superior convergence speed. As evident in the visualizations, YCbCr-based reconstruction shows significant improvements within just 10 iterations, achieving sharper results compared to RGB. This acceleration in convergence can be attributed to the reduced dimensionality of color information: YCbCr isolates intensity variations to the Y channel, with only two channels handling color information. This characteristic appears to enhance optimization stability by reducing inter-channel interference during the reconstruction process.

\begin{table*}
\centering
\resizebox{0.98\linewidth}{!}{%
\setlength{\tabcolsep}{2pt}
\begin{tabular}{c*{8}{c}}
\toprule
CelebA & & ITER=0 & 100 & 500 & 1000 & 5000 & 10000 \\
\midrule
\multirow{3}{*}{\parbox{4.5cm}{\small\centering GaussianImage}} 
& PSNR & - & $10.3\pm2.4$ & $21.9\pm1.2$ & $30.0\pm2.1$ & $43.8\pm2.9$ & $45.2\pm2.8$ \\
& SSIM & - & $0.47\pm0.05$ & $0.82\pm0.04$ & $0.95\pm0.02$ & $0.99\pm0.001$ & $0.99\pm0.001$ \\
& \% & - & $0$ & $0$ & $0$ & $91$ & $97$ \\
\midrule
\multirow{3}{*}{\parbox{4.5cm}{\small\centering Ours (CelebA-trained EigenGS)}} 
& PSNR & $29.9$ & $37.8\pm3.8$ & $41.6\pm3.9$ & $43.6\pm3.6$ & $47.1\pm3.1$ & $48.1\pm3.1$ \\
& SSIM & $0.93$ & $0.98\pm0.02$ & $0.99\pm0.01$ & $0.99\pm0.003$ & $0.99\pm0.001$ & $0.99\pm0.001$\\
& \% & - & $28$ & $66$ & $89$ & $98$ & $99$ \\
\midrule
\multirow{3}{*}{\parbox{4.5cm}{\small\centering Ours (ImageNet-trained EigenGS)}} 
& PSNR & $28.7$ & $35.4\pm3.4$ & $39.6\pm3.7$ & $42.2\pm3.6$ & $46.3\pm3.2$ & $47.3\pm3.1$ \\
& SSIM & $0.91$ & $0.96\pm0.03$ & $0.99\pm0.01$ & $0.99\pm0.01$ & $0.99\pm0.002$ & $0.99\pm0.001$\\
& \% & - & $10$ & $47$ & $80$ & $98$ & $98$ \\
\bottomrule
\end{tabular}
}
\caption{Quantitative comparison on the CelebA dataset with PSNR, SSIM, and percentage of images achieving PSNR larger than 40 dB.}
\label{tab:suppl_celeba}
\end{table*}

\begin{table*}
\centering
\resizebox{0.98\linewidth}{!}{%
\setlength{\tabcolsep}{2pt}
\begin{tabular}{c*{8}{c}}
\toprule
FFHQ & & ITER=0 & 100 & 500 & 1000 & 5000 & 10000 \\
\midrule
\multirow{3}{*}{\parbox{4.5cm}{\small\centering GaussianImage}} 
& PSNR & - & $10.4\pm1.7$ & $21.8\pm0.9$ & $29.4\pm1.6$ & $39.2\pm1.9$ & $40.1\pm1.9$ \\
& SSIM & - & $0.41\pm0.05$ & $0.77\pm0.05$ & $0.94\pm0.03$ & $0.99\pm0.001$ & $0.99\pm0.001$ \\
& \% & - & $0$ & $0$ & $0$ & $98$ & $99$ \\
\midrule
\multirow{3}{*}{\parbox{4.5cm}{\small\centering Ours (FFHQ-trained EigenGS)}} 
& PSNR & $28.0$ & $34.4\pm2.4$ & $36.4\pm2.6$ & $37.5\pm2.6$ & $40.7\pm2.5$ & $41.8\pm2.4$ \\
& SSIM & $0.87$ & $0.95\pm0.02$ & $0.98\pm0.01$ & $0.99\pm0.01$ & $0.99\pm0.003$ & $0.99\pm0.002$\\
& \% & - & $41$ & $76$ & $83$ & $98$ & $99$ \\
\midrule
\multirow{3}{*}{\parbox{4.5cm}{\small\centering Ours (ImageNet-trained EigenGS)}} 
& PSNR & $27.2$ & $34.1\pm2.5$ & $36.1\pm2.6$ & $37.2\pm2.7$ & $40.5\pm2.6$ & $41.6\pm2.4$ \\
& SSIM & $0.84$ & $0.95\pm0.02$ & $0.97\pm0.01$ & $0.98\pm0.01$ & $0.99\pm0.003$ & $0.99\pm0.002$\\
& \% & - & $39$ & $71$ & $79$ & $97$ & $99$ \\
\bottomrule
\end{tabular}
}
\caption{Quantitative comparison on the FFHQ dataset with PSNR, SSIM, and percentage of images achieving PSNR larger than 35 dB.}
\label{tab:suppl_ffhq}
\end{table*}

\begin{table*}
\centering
\resizebox{0.98\linewidth}{!}{%
\setlength{\tabcolsep}{2pt}
\begin{tabular}{c*{8}{c}}
\toprule
Cats & & ITER=0 & 100 & 500 & 1000 & 5000 & 10000 \\
\midrule
\multirow{3}{*}{\parbox{4.5cm}{\small\centering GaussianImage}} 
& PSNR & - & $11.2\pm2.2$ & $22.2\pm1.3$ & $30.4\pm2.2$ & $42.4\pm4.9$ & $43.2\pm4.9$ \\
& SSIM & - & $0.47\pm0.10$ & $0.82\pm0.07$ & $0.96\pm0.02$ & $0.99\pm0.02$ & $0.99\pm0.02$ \\
& \% & - & $0$ & $0$ & $0$ & $69$ & $74$ \\
\midrule
\multirow{3}{*}{\parbox{4.5cm}{\small\centering Ours (Cats-trained EigenGS)}} 
& PSNR & $30.6$ & $38.1\pm4.8$ & $41.3\pm5.2$ & $42.8\pm5.1$ & $45.3\pm4.5$ & $46.1\pm4.5$ \\
& SSIM & $0.92$ & $0.97\pm0.02$ & $0.99\pm0.01$ & $0.99\pm0.01$ & $0.99\pm0.001$ & $0.99\pm0.001$\\
& \% & - & $35$ & $57$ & $70$ & $89$ & $90$ \\
\midrule
\multirow{3}{*}{\parbox{4.5cm}{\small\centering Ours (ImageNet-trained EigenGS)}} 
& PSNR & $29.6$ & $37.6\pm5.2$ & $41.1\pm5.6$ & $42.9\pm5.4$ & $45.9\pm4.7$ & $46.6\pm4.6$ \\
& SSIM & $0.90$ & $0.96\pm0.04$ & $0.99\pm0.02$ & $0.99\pm0.01$ & $0.99\pm0.002$ & $0.99\pm0.002$\\
& \% & - & $35$ & $59$ & $67$ & $88$ & $90$ \\
\bottomrule
\end{tabular}
}
\caption{Quantitative comparison on the Cats dataset with PSNR, SSIM, and percentage of images achieving PSNR larger than 40 dB.}
\label{tab:suppl_cats}
\end{table*}

\begin{table*}
\centering
\resizebox{0.98\linewidth}{!}{%
\setlength{\tabcolsep}{2pt}
\begin{tabular}{c*{8}{c}}
\toprule
Cars & & ITER=0 & 100 & 500 & 1000 & 5000 & 10000 \\
\midrule
\multirow{3}{*}{\parbox{4.5cm}{\small\centering GaussianImage}} 
& PSNR & - & $12.7\pm0.8$ & $23.9\pm0.7$ & $32.9\pm1.6$ & $41.8\pm2.3$ & $42.9\pm2.2$ \\
& SSIM & - & $0.57\pm0.05$ & $0.03\pm0.01$ & $0.98\pm0.001$ & $0.99\pm0.004$ & $0.99\pm0.001$ \\
& \% & - & $0$ & $0$ & $10$ & $100$ & $100$ \\
\midrule
\multirow{3}{*}{\parbox{4.5cm}{\small\centering Ours (Cars-trained EigenGS)}} 
& PSNR & $24.4$ & $31.6\pm2.5$ & $34.1\pm3.1$ & $36.6\pm3.6$ & $43.3\pm4.2$ & $44.7\pm4.1$ \\
& SSIM & $0.85$ & $0.95\pm0.02$ & $0.98\pm0.01$ & $0.99\pm0.01$ & $0.99\pm0.001$ & $0.99\pm0.001$\\
& \% & - & $11$ & $32$ & $67$ & $99$ & $100$ \\
\midrule
\multirow{3}{*}{\parbox{4.5cm}{\small\centering Ours (ImageNet-trained EigenGS)}} 
& PSNR & $23.9$ & $31.5\pm2.5$ & $33.9\pm2.9$ & $36.3\pm3.2$ & $43.2\pm4.1$ & $44.5\pm4.1$ \\
& SSIM & $0.84$ & $0.95\pm0.03$ & $0.98\pm0.01$ & $0.99\pm0.01$ & $0.99\pm0.02$ & $0.99\pm0.01$\\
& \% & - & $8$ & $38$ & $62$ & $98$ & $99$ \\
\bottomrule
\end{tabular}
}
\caption{Quantitative comparison on the Cars dataset with PSNR, SSIM, and percentage of images achieving PSNR larger than 35 dB.}
\label{tab:suppl_cars}
\end{table*}

\begin{figure*}[t]
    \centering
    
    \begin{minipage}{0.19\textwidth}
        \centering
        ITER = 0
    \end{minipage}%
    \begin{minipage}{0.19\textwidth}
        \centering
        ITER = 10
    \end{minipage}%
    \begin{minipage}{0.19\textwidth}
        \centering
        ITER = 100
    \end{minipage}
    \begin{minipage}{0.19\textwidth}
        \centering
        ITER = 1000
    \end{minipage}
    \begin{minipage}{0.19\textwidth}
        \centering
        ITER = 10000
    \end{minipage}
    \vspace{0.3em}

    \begin{subfigure}[b]{0.19\textwidth}
        \includegraphics[width=\textwidth]{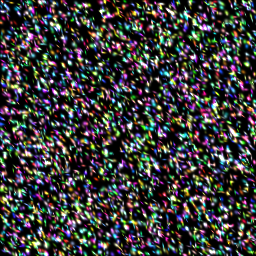}
    \end{subfigure}
    \begin{subfigure}[b]{0.19\textwidth}
        \includegraphics[width=\textwidth]{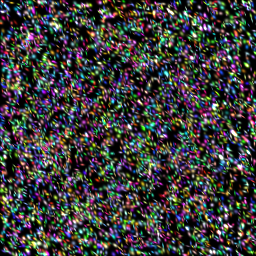}
    \end{subfigure}
    \begin{subfigure}[b]{0.19\textwidth}
        \includegraphics[width=\textwidth]{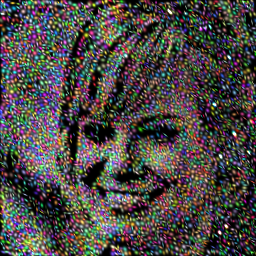}
    \end{subfigure}
    \begin{subfigure}[b]{0.19\textwidth}
        \includegraphics[width=\textwidth]{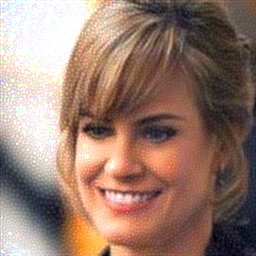}
    \end{subfigure}
    \begin{subfigure}[b]{0.19\textwidth}
        \includegraphics[width=\textwidth]{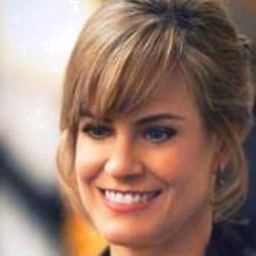}
    \end{subfigure}

    \begin{subfigure}[b]{0.19\textwidth}
        \includegraphics[width=\textwidth]{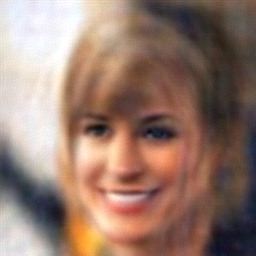}
    \end{subfigure}
    \begin{subfigure}[b]{0.19\textwidth}
        \includegraphics[width=\textwidth]{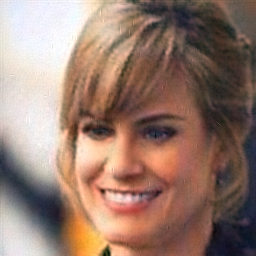}
    \end{subfigure}
    \begin{subfigure}[b]{0.19\textwidth}
        \includegraphics[width=\textwidth]{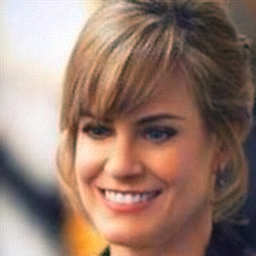}
    \end{subfigure}
    \begin{subfigure}[b]{0.19\textwidth}
        \includegraphics[width=\textwidth]{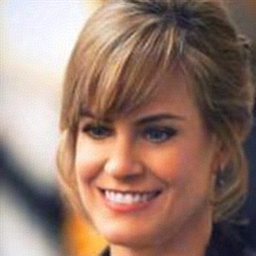}
    \end{subfigure}
    \begin{subfigure}[b]{0.19\textwidth}
        \includegraphics[width=\textwidth]{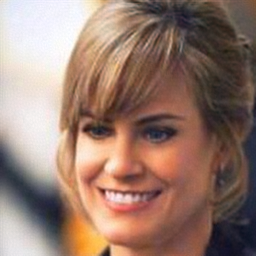}
    \end{subfigure}

    \caption{Visual comparison between GaussianImage (top) and our method (bottom).}
    \label{fig:vis_baseline}
\end{figure*}

\end{document}